\documentclass[onecolumn]{IEEEtran}
\usepackage{amsfonts, amssymb, amsmath}
\usepackage{epsfig}
\usepackage{theorem}
\usepackage{textcomp}
\usepackage{multirow}
\usepackage{tabularx}
\usepackage{graphicx}
\usepackage{epstopdf}

\makeatletter
\let\NAT@parse\undefined
\makeatletter
\usepackage{natbib}

\usepackage{xr}
\newtheorem{theorem}{Theorem}

\newtheorem{proposition}[theorem]{Proposition}
\newtheorem{corollary}[theorem]{Corollary}

\newtheorem{problem}{Problem}
\newtheorem{objective}{Objective}

\def\remark{\noindent\textbf{Remark. }}

\newcommand{\qed}{\hfill $\Box$\\}
\newcommand{\rqed}{\hfill $\triangle$\\}

\def\tompp{{\sc MinMakespan}}
\def\mdmpp{{\sc MinMaxDist}}
\def\dompp{{\sc MinTotalDist}}
\def\ttmpp{{\sc MinTotalTime}}
\def\mpp{\textrm{MPP}}

\def\odid{{\sc OD+ID}}
\def\ida{{\sc ID}}
\def\wcha{{\sc WHCA*}}
\def\cobopt{{\sc Cobopt}}

\begin{document}

\title{Optimal Multi-Robot Path Planning on Graphs: Complete Algorithms and Effective Heuristics}
\author{
\begin{tabular}{ccc}
Jingjin Yu & & Steven M. LaValle 
\end{tabular}
\thanks{
Jingjin Yu is with the Computer Science and Artificial Intelligence Lab, Massachusetts Institute of Technology, Cambridge, MA 02139, USA. E-mail: jingjin@csail.mit.edu. Steven M. LaValle is with the Department of Computer Science, University of Illinois at Urbana-Champaign, Urbana, IL 61801 USA. E-mail: lavalle@illinois.edu.
%This work was supported in part by NSF grants 0904501 (IIS Robotics) and 1035345 (Cyberphysical Systems), DARPA SToMP grant HR0011-05-1-0008, and MURI/ONR grant N00014-09-1-1052. Jingjin Yu is with the Department of Electrical and Computer Engineering, University of Illinois at Urbana-Champaign, Urbana, IL 61801 USA. E-mail: jyu18@uiuc.edu. Steven M. LaValle is with the Department of Computer Science, University of Illinois at Urbana-Champaign, Urbana, IL 61801 USA. E-mail: lavalle@uiuc.edu.
}
}
\maketitle

\begin{abstract}We study the problem of optimal multi-robot path planning on graphs ($\mpp$) over four distinct minimization objectives: the makespan (last arrival time), the maximum (single-robot traveled) distance, the total arrival time, and the total distance. In a related paper \cite{YuLav15TRO-I}, we show that these objectives are distinct and NP-hard to optimize. In this work, we focus on efficiently algorithmic solutions for solving these optimal $\mpp$ problems. Toward this goal, we first establish a one-to-one solution mapping between $\mpp$ and network-flow. Based on this equivalence and integer linear programming (ILP), we design novel and complete algorithms for optimizing over each of the four objectives. In particular, our exact algorithm for computing optimal makespan solutions is a first such that is capable of solving extremely challenging problems with robot-vertex ratio as high as $100\%$. Then, we further improve the computational performance of these exact algorithms through the introduction of principled heuristics, at the expense of some optimality loss. The combination of ILP model based algorithms and the heuristics proves to be highly effective, allowing the computation of $1.x$-optimal solutions for problems containing hundreds of robots, densely populated in the environment, often in just seconds. 
\end{abstract}

\section{Introduction}

\label{sec:intro}
In this paper, we study the problem of optimal {\em multi-robot path planning on graphs} ($\mpp$), focusing on the design of {\em complete algorithms and effective heuristics}. In an $\mpp$ instance, the robots are uniquely labeled ({\em i.e.}, distinguishable) and are confined to an arbitrary connected graph. The robots may move from a vertex to an adjacent one in one time step in the absence of collision, which may occur when two robots simultaneously move to the same vertex or along the same edge in different directions. A distinguishing feature of our $\mpp$ formulation is that we allow robots on fully occupied cycles to rotate synchronously. Such a formulation, more appropriate for multi-robot applications, has not been widely studied (except, {\em e.g.}, \cite{Sta10,StaKor11}). Over the basic $\mpp$ formulation, we look at four commonly studied minimization objectives: the makespan (last arrival time), the maximum (single-robot traveled) distance, the total arrival time, and the total distance. We note that these global objectives have direct relevance toward real world multi-robot applications such as autonomous warehouse systems \cite{WurDanMou08}. For example, minimizing makespan is equivalent to minimizing the task completion time, whereas minimizing total distance is applicable to minimizing the fuel consumption of the entire fleet of robots. 

In a related paper \cite{YuLav15TRO-I}, we show that these objectives are all distinct and NP-hard to optimize, suggesting that efforts on solving optimal $\mpp$ should be directed at finding effective, near-optimal algorithms. In this work, we make an attempt toward this goal and propose a novel yet general framework for solving $\mpp$ optimally. By examining space and time dimensions jointly, we observe a one-to-one mapping between a solution for an $\mpp$ instance and that for a multi-commodity flow problem derived from the $\mpp$ problem. Based on the equivalence relationship, we can then translate the $\mpp$ problem into an integer linear programming (ILP) model, which can be subsequently solved using an ILP solver. The generality of ILP allows the encoding of all four objectives to yield complete algorithms for optimizing these objectives. From here, we take a further step and introduce several heuristics to boost the algorithmic performance at a slight loss of solution optimality. Our method is especially effective in computing near-optimal minimum makespan solutions, capable of computing $1.x$-optimal solutions for hundreds of robots, densely populated on the underlying graph, often in just seconds. 

{\bf Related work.} Multi-robot path planning problems, in its many formulations, have been actively studied for decades \cite{ErdLoz86,Zel92,LavHut98b,Sil05,KloHut06,Rya08,JanStu08,Sur09,LunBer11,StaKor11,BerOve05, BerSnoLinMan09, SolHal12,YuLav13STAR,TurMicKum14,SolYu15}. As a universal subroutine, collision-free path planning for multiple robots finds applications in tasks spanning assembly \cite{HalLatWil00, Nna92}, evacuation \cite{RodAma10}, formation control \cite{BalArk98, PodSuk04, ShuMurBen07,  SmiEgeHow08, TanPapKum04}, localization \cite{FoxBurKruThr00}, micro droplet manipulation \cite{DinChaFai01,GriAke05}, object transportation \cite{MatNilSim95, RusDonJen95}, search and rescue \cite{JenWheEva97}, and so on. We refer the readers to \cite{ChoLynHutKanBurKavThr05, Lat91, Lav06} and the references therein for a more comprehensive review on the general subject of multi-robot path and motion planning. 

The algorithmic study of graph-based multi-robot path planning problems, the focus of this work, can be traced to as least 1879 \cite{Sto1879}, in which Story makes the observation that the feasibility of the $15$-puzzle \cite{Loy59} is decided by the parity of the game. The $15$-puzzle is a restricted $\mpp$ instance moving $15$ labeled game pieces on a $4\times 4$ grid, from some initial configuration to some goal configuration. The restriction is that only a single game piece near the only empty vertex may move to the empty vertex in a step--multiple mobile robots could move simultaneously. A generalization of the $15$-puzzle is introduced in \cite{Wil74}, extending the problem from $15$ game pieces on a $4\times 4$ grid to $n - 1$ labeled {\em pebbles} on an $n$-vertex, $2$-connected graph. It is shown, together with an implied algorithm, that an instance is always feasible if the graph is non-bipartite. When the graph is bipartite (such as the $15$-puzzle), all pebble configurations are split into two groups of equal size such that any two configurations in the same group form a feasible instance. A further generalization is introduced in \cite{KorMilSpi84}, allowing $p < n$ pebbles on a graph with $n$ vertices. For this problem, an $O(n^3)$ algorithm is provided to solve an instance or decide that the instance is infeasible. 

As computer games and multi-robot systems gain popularity, concurrent movements are introduced and pebbles are replaced with robots (or agents). On the feasibility side, the $\mpp$ problem studied in this paper is shown to be solvable also in $O(n^3)$ time \cite{YuRus14WAFR}. To distinguish the formulations, we denote the formulation that does not allow cyclic rotations of robots along fully occupied cycles as {\em cycle-free} $\mpp$. Until recently, the majority of algorithmic study on $\mpp$ is on the cycle-free case. Since the problem is shown to be tractable \cite{KorMilSpi84}, most algorithmic study of cycle-free $\mpp$ put some emphasis on optimality. Through the clever use of primitive operations, algorithms from \cite{Rya08,Sur09,LunBer11,SajLunBek12,WilMorWit14} could quickly solve difficult problems with some form of completeness guarantees. These algorithms do not have optimality guarantees but the produced solutions are often of much better quality than the $O(n^3)$ bound given by \cite{KorMilSpi84}. For more discussion and references on sub-optimal methods, see \cite{SajLunBek12,WilMorWit14}. 

Pushing more toward the optimality side, most algorithmic results explore ways to limit the exponential search space growth induced by multiple robots. One of the first such algorithms, Local Repair A* (LRA*) \cite{Zel92}, plans robot paths simultaneously and performs local repairs when conflicts arise. Focusing on fixing the (locality) shortcomings of LRA*, Windowed Hierarchical Cooperative A* (\wcha) proposes using a space-time window to allow more choices for resolving local conflicts while simultaneously limiting the search space size \cite{Sil05}. Iterative deepening technique is shown to be applicable to $\mpp$ problems in \cite{ShaSteFelStu12}. A technique called sub-dimensional expansion is shown to perform well in complex environment \cite{WagChoC11}; the robot density is however relatively low ($104$ cells per robot according to the paper). In \cite{Sta10,StaKor11}, instead of agnostic dissection of an instance, the natural idea of independence detection (ID) is explored to only consider multiple robots jointly (the source of exponential search space growth) as necessary. With operator decomposition (OD) that treats each legal move as an ``operator'', the authors produced algorithms (\ida,\odid, and related variants) that prove to be quite effective in computing total time- or distance-optimal solutions. We point out that \ida\, and \odid\, have support for handling cycles ({\em i.e.}, they apply to $\mpp$ instead of cycle-free $\mpp$). Algorithms designed for minimizing makespan have also been attempted, {\em e.g.}, \cite{Sur12}, but the solution quality degrades rapidly as the robot-vertex ratio increases.

Many approaches have also been proposed for solving multi-robot path planning problems in the continuous domain. A representative method called velocity-obstacles \cite{KanZuc86,blm-rvo,BerSnaGuyMan11} explicitly examines velocity-time space for coordinating robot motions. In \cite{GriAke05}, mixed integer programming (MIP) models are employed to encode the robot interactions. A method based on the space-time perspective, similar to ours, is explored in \cite{KarGerSta12}. In \cite{PeaClaMcp08}, an A*-based search is performed over a discrete roadmap abstracted from the continuous environment. In \cite{SolSalHal14}, discrete-RRT (d-RRT) is proposed for the efficient search of multi-robot roadmaps. Algorithms for discrete $\mpp$, cycle-free or not, have also helped solving continuous problems \cite{SolHal12,ksb-tdmp12}. 

{\bf Contributions.} We study the optimal $\mpp$ formulation allowing up to $n$ robots on a $n$-vertex connected graph, which we believe is better suited for multi-robot applications. The formulation is not widely studied, perhaps due to the inherent difficulty in handling cyclic rotations of robots. Besides the novelty of the problem, this work brings several algorithmic contributions. First, based on the equivalence relationship between $\mpp$ and network flow, we establish a general and novel solution framework allowing the compact encoding of optimal $\mpp$ problems using ILP models. We show that the framework readily produces complete algorithms for minimizing the makespan (last arrival time), the maximum (single-robot traveled) distance, the total arrival time, and the total distance, which are perhaps four most common global objectives for $\mpp$. The resulting algorithms, in particular the one for computing minimum makespan, are highly effective in computing challenging problems instances with robot-vertex ratio up to $100\%$. Second, we introduce several principled heuristics, in particular a $k$-way split heuristic that divides an $\mpp$ instance over the time horizon, to give the exact algorithms a sizable performance boost at the expense of some solution optimality loss. With these heuristics, we are able to extend our algorithms to tackle problems with several hundred robots that are extremely densely populated, while at the same time maintain $1.x$ solution optimality. Last but not least, our successful exploit of ILP to attack optimal $\mpp$ shows that integer linear programming method is competitive with direct search methods in this problem domain, especially when the number of robots becomes large. This is surprising because ILP solvers are not designed specifically for $\mpp$. 

The rest of the paper is organized as follows. In Section \ref{sec:formulation}, we define the optimal $\mpp$ problems and provide background material on network flow. We then establish the equivalence relationship between $\mpp$ and multi-commodity flow in Section~\ref{sec:flow}. We derive complete algorithms in Section~\ref{sec:model} based on the equivalence and continue to describe the performance boosting heuristics in Section~\ref{section:heuristics}. We evaluate the algorithms in Section~\ref{sec:evaluation} and conclude in Section~\ref{sec:conclusion}. This paper is partly based on \cite{YuLav13STAR,YuLav13ICRA-A,YuLav13AAAI-LBP}.\footnote{\cite{YuLav13AAAI-LBP} is a preliminary poster presentation.} In comparison to \cite{YuLav13STAR,YuLav13ICRA-A}, besides demonstrating significantly improved computational performance due to the addition of the $k$-way split heuristic, we have substantially extended the generality of our ILP-based algorithmic framework, which now supports all common global time- and distance-based objectives. 

\section{Preliminaries}\label{sec:formulation}
We now define $\mpp$ and the optimality objectives studied in this paper. Following the problem statements, we gave a brief review on {\em network flow}. 

\subsection{Multi-Robot Path Planning on Graphs}
\begin{figure}[htp]
\begin{center}
  \begin{tabular}{ccc}
    \includegraphics[width=0.12\textwidth]{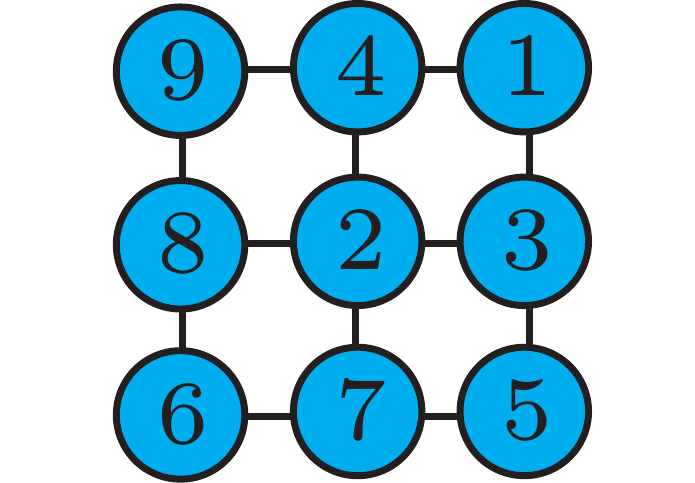} & &
    \includegraphics[width=0.12\textwidth]{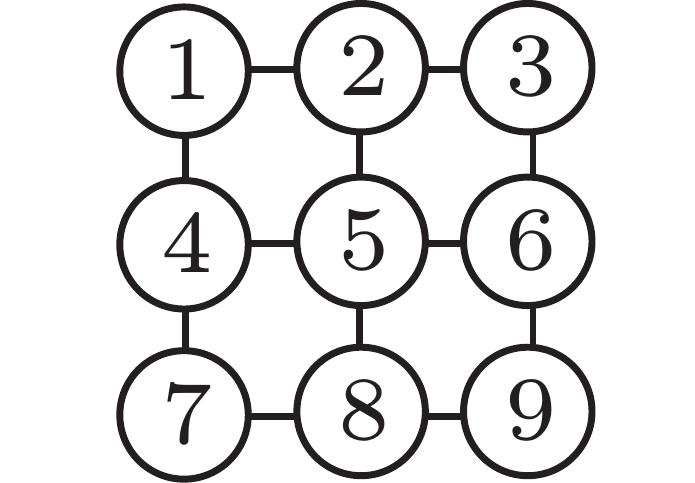}  \\
    (a) && (b)\\
  \end{tabular}
\end{center}
\caption{\label{fig:example} a) A 9-puzzle problem. b) The desired goal configuration.}
\end{figure}

Let $G = (V, E)$ be a connected, undirected, simple graph, with $V = \{v_i\}$ being the vertex set and $E = \{(v_i, v_j)\}$ the edge set. Let $R = \{r_1, \ldots, r_n\}$ be a set of $n$ robots. The robots move at discrete time steps ({\em i.e.}, at $t = 0, 1, \ldots$). At time step $t = 0$, each robot occupies a distinct vertex of $G$. In general, at any time step $t = 0, 1, \ldots$, the robots assume a {\em configuration} that is an injective map from $R$ to $V$. The {\em start} ({\em initial}) and {\em goal} configurations of the robots are denoted as $x_I$ and $x_G$, respectively. Fig.~\ref{fig:example}(a) shows a possible configuration of $9$ robots on a $3\times 3$ grid graph. Fig.~\ref{fig:example}(b) shows a possible goal configuration, in which the robots are ordered based on {\em row-major ordering} \footnote{In this paper, we generally use shaded discs to mark start locations of robots and discs without shading for goal locations.}.

\begin{figure}[ht!]
\begin{center}
  \begin{tabular}{ccc}
    \includegraphics[width=0.12\textwidth]{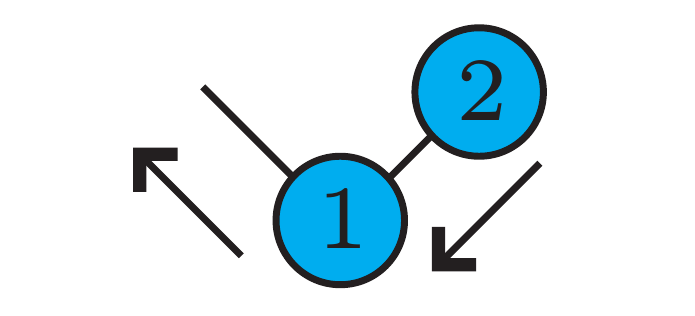} & 
    \includegraphics[width=0.09\textwidth]{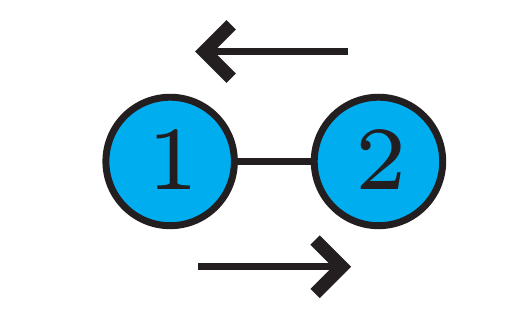} & 
    \includegraphics[width=0.09\textwidth]{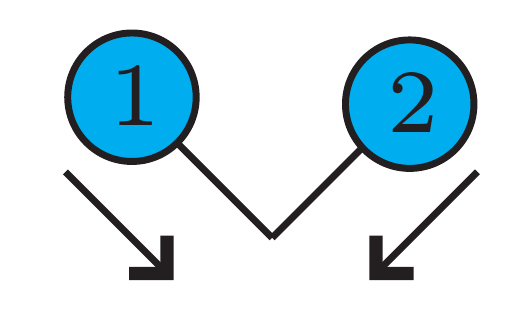}  \\
    (a) & (b) & (c)\\
  \end{tabular}
\end{center}
\caption{\label{fig:moves} Some feasible and infeasible moves for two robots. a) A feasible synchronous move. b) An infeasible synchronous move in which two robot collide ``head-on''. c) An infeasible synchronous move in which two robots ``meet'' at a vertex.}
\end{figure}

During a discrete time step, each robot may either remain stationary or move to an adjacent vertex. To formally describe a plan, let a {\em scheduled path} be a map $p_i: \mathbb Z^+ \to V$, in which $\mathbb Z^+ := \mathbb N \cup \{0\}$. A scheduled path $p_i$ is {\em feasible} if it satisfies the following properties: 1) $p_i(0) = x_I(r_i)$. 2) For each $i$, there exists a smallest $t_i \in \mathbb Z^+$ such that $p_i(t_i) = x_G(r_i)$. 3) For any $t \ge t_i$, $p_i(t) \equiv x_G(r_i)$. 4) For any $0 \le t < t_i$, $(p_i(t), p_i(t+1)) \in E$ or $p_i(t) = p_i(t+1)$ (if $p_i(t) = p_i(t+1)$, robot $r_i$ stays at vertex $p_i(t)$ between the time steps $t$ and $t+1$). We say that two paths $p_i, p_{j}$ are in {\em collision} if there exists $k \in \mathbb Z^+$ such that $p_i(t) = p_{j}(t)$ (meet collision) or $(p_i(t), p_i(t+1)) = (p_j(t+1), p_j(t))$ (head-on collision). As an illustration, Fig.~\ref{fig:moves} shows the feasible and infeasible moves for two robots during a single time step \footnote{We assume that the graph $G$ only allows ``meet'' or ``head-on'' collisions. The assumption is mild. For example, a (arbitrary dimensional) grid with unit edge distance is such a graph for robots of with radii of no more than $\sqrt{2}/4$.}. The {\em multi-robot path planning on graph} (\mpp) problem is defined as follows. 

\begin{problem}[Multi-robot Path Planning on Graphs]\label{mpp} Given a $4$-tuple $(G, R, x_I, x_G)$, find a set of paths $P = \{p_1, \ldots, p_n\}$ such that $p_i$'s are feasible paths for respective robots $r_i$'s and no two paths $p_i, p_j$ are in collision. 
\end{problem}

For example, Fig.~\ref{fig:example}(a) and Fig.~\ref{fig:example}(b) define an $\mpp$ problem on the $3\times 3$ grid. We call this particular problem the {\em 9-puzzle} problem (a variant of the 15-puzzle \cite{RatWar90}), which readily generalizes to $N^2$-puzzles. 
\vspace{2mm}

\remark With a few exceptions ({\em e.g.}, \cite{StaKor11}), most existing studies on discrete multi-robot path planning problems require empty vertices as swap spaces. In these formulations, in a time step, a non-intersecting chain of robots may move simultaneously only if the head of the chain is moving into a previously unoccupied vertex. In contrast, our $\mpp$ formulation allows synchronized rotations of robots along fully occupied cycles (see, {\em e.g.}, Fig. \ref{fig:example} and \ref{fig:puzzle-8-sol}). This implies that even when the number of robots equals the number of vertices, robots can still move on disjoint cycles. We note that $\mpp$ can be solved in polynomial time with feasibility test taking only linear time~\cite{YuRus14WAFR}. ~\rqed

\subsection{Optimal Formulations}
Let $P = \{p_1, \ldots, p_n\}$ be an arbitrary feasible solution to some fixed $\mpp$ instance. For a path $p_i \in P$, let $len(p_i)$ denote the length of the path $p_i$, which is increased by one each time when the robot $r_i$ passes an edge. A robot, following a path $p_i$, may visit the same edge multiple times. Recall that $t_i$ denotes the arrival time of robot $r_i$. In the study of optimal $\mpp$ formulations, we examine four common objectives with two focusing on time optimality and two focusing on distance optimality. Below, each objective is formally defined.

\begin{objective}[Makespan]\label{omakespan} Compute a path set $P$ that minimizes $\max_{1 \le i \le n}t_i$.
%\begin{displaymath}
%\max_{1 \le i \le n}t_i.
%\end{displaymath}
\end{objective}

\begin{objective}[Maximum Distance]\label{md}Compute a path set $P$ that minimizes $\max_{1 \le i \le n} len(p_i)$. 
%\begin{displaymath}
%\max_{1 \le i \le n}len(p_i).
%\end{displaymath}
\end{objective}

\begin{objective}[Total Arrival Time]\label{ott} Compute a path set $P$ that minimizes $\sum_{i = 1}^nt_i$.
%\begin{displaymath}
%\sum_{i = 1}^nt_i.
%\end{displaymath}
\end{objective}

\begin{objective}[Total Distance]\label{otd}Compute a path set $P$ that minimizes $\sum_{i = 1}^n len(p_i)$. 
%\begin{displaymath}
%\sum_{i = 1}^n len(p_i).
%\end{displaymath}
\end{objective}

\begin{figure}[htp]
\begin{center}
    \includegraphics[width=0.48\textwidth]{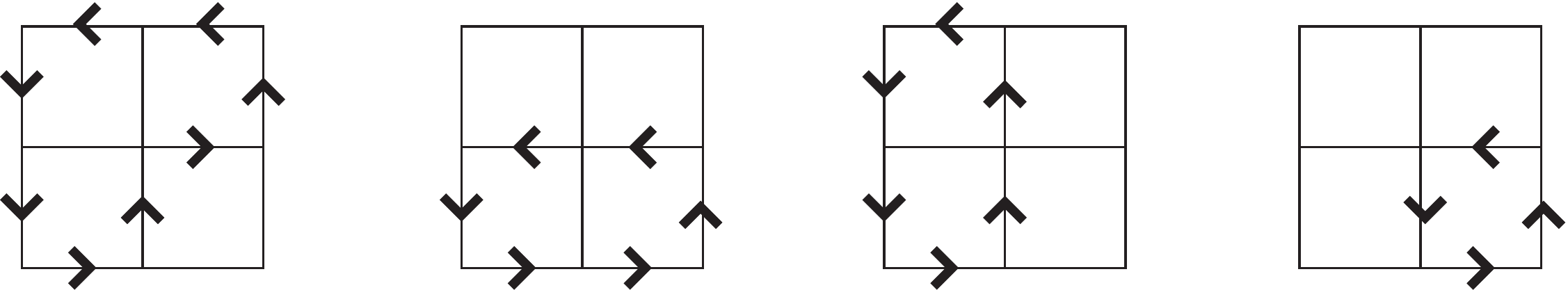} 
\end{center}
\caption{\label{fig:puzzle-8-sol} A 4-step solution from our algorithm. The directed edges show the moving directions of the robots at the tail of the edges.}
\end{figure}

The intuitive meaning of these objectives is clear. Fig.~\ref{fig:puzzle-8-sol} illustrates the four-step minimum makespan solution to the 9-puzzle problem from Fig.~\ref{fig:example}. The solution is optimal as it takes at least four steps for robot $9$ to move to its goal.

\subsection{Network Flow Review}\label{subsection:network-flow-review}
A {\em network} $\mathcal N = (G, c_1, c_2, S)$ consists of a {\em directed graph} $G = (V, E)$ with $c_1, c_2: E \to \mathbb Z^+$ as the maps specifying {\em capacities} and {\em costs} over edges, respectively, and $S \subset V$ as the set of {\em sources} and {\em sinks}. Let $S = S^+ \cup S^-$, with $S^+$ denoting the set of source vertices, $S^-$ denoting the set of sink vertices, and $S^+ \cap S^- = \varnothing$. For a vertex $v \in V$, let $\delta^+(v)$ (resp., $\delta^-(v)$) denote the set of edges of $G$ going to (resp., leaving) $v$. A feasible (static) $S^+, S^-$-flow on this network $\mathcal N$ is a map $f: E \to \mathbb Z^+$ that satisfies edge capacity constraints,
\begin{equation}\label{c1}
\forall e \in E, \quad f(e) \le c_1(e),
\end{equation}
the flow conservation constraints at non terminal vertices,
\begin{equation}\label{c2}
\forall v \in V \backslash S, \quad \displaystyle\sum_{e\in \delta^+(v)} f(e)\,\,\, - \sum_{e\in \delta^-(v)} f(e) = 0,
\end{equation}
and the flow conservation constraints at terminal vertices,
\begin{equation}\label{flow-value}
\begin{array}{ll}
F(f) &= \displaystyle\sum_{v \in S^+} (\sum_{e\in \delta^-(v)} f(e)\,\,\, - \sum_{e\in \delta^+(v)} f(e)) \\ 
& = \displaystyle\sum_{v \in S^-} (\sum_{e\in \delta^+(v)} f(e)\,\,\, - \sum_{e\in \delta^-(v)}f(e)).
\end{array}
\end{equation}
The quantity $F(f)$ is called the {\em value} of the flow $f$. The classic (single-commodity) {\em maximum flow} problem asks the following question: Given a network $\mathcal N$, what is the maximum $F(f)$ that can be pushed through the network? The {\em minimum cost maximum flow} problem further requires the flow to have a minimum total cost among all maximum flows. That is, we want to find a flow among all maximum flows that also minimizes the quantity
\begin{equation}\label{min-cost-max-flow}
\sum_{e \in E} c_2(e)\cdot f(e).
\end{equation}

The network flow formulation described so far only considers a {\em single commodity}, corresponding to all robots being interchangeable. For general $\mpp$ formulations, the robots are distinct and must be treated as different commodities. Such problems can be captured with the {\em Multi-commodity flow} or simply {\em multiflow}. Instead of having a single flow function $f$, we have a flow function $f_i$ for each commodity $i$. The constraints~\eqref{c1},~\eqref{c2}, and~\ref{flow-value} become
\begin{equation}\label{c1m}
\forall i, \forall e \in E, \quad \sum_i \,\, f_i(e) \le c_1(e),
\end{equation}
\begin{equation}\label{c2m}
\forall \, i, \forall \, v \in V \backslash S, \quad \displaystyle\sum_{e\in \delta^+(v)} f_i(e)\,\,\, - \sum_{e\in \delta^-(v)} f_i(e) = 0,
\end{equation}
\begin{equation}\label{flow-value-m}
\begin{array}{lll}
\forall i, & &\displaystyle\sum_{v \in S^+} (\sum_{e\in \delta^-(v)} f_i(e)\,\,\, - \sum_{e\in \delta^+(v)} f_i(e)) \\ 
&=& \displaystyle\sum_{v \in S^-} (\sum_{e\in \delta^+(v)} f_i(e)\,\,\, - \sum_{e\in \delta^-(v)}f_i(e)).
\end{array}
\end{equation}

Maximum flow and minimum cost flow problems may also be posed under a multiflow setup; we omit the details. Our review on network flow only touches aspects pertinent to the current work; for a thorough coverage on the subject of network flow, see \cite{Aro89,AhuMagOrl93} and the references therein. Note that the multiflow model stated here is sometimes also referred to as {\em integer multiflow} because $f_i$ must have integer values.

\section{From Multi-Robot Path Planning to Multiflow}\label{sec:flow}
A close algorithmic connection exists between optimal $\mpp$ and network flow problems. Maximum (single commodity) flow problems generally admit efficient (low-degree polynomial time) algorithmic solutions \cite{AhuMagOrl93}, whereas maximum multiflow is a well-known NP-hard problem, difficult to even approximate \cite{AndZha05}. Mirroring the disparity between single- and multi-commodity flows, in the domain of $\mpp$ problems, if there is a single group of interchangeable robots (here, for a group, it does not matter which robot goes to which goal as long as all goal locations assigned to the group are occupied by robots from the same group), then many optimal formulations admit polynomial time algorithms \cite{YuLav13STAR}. However, as soon as a single group of robots splits into two or more groups, finding optimal paths for these robots become intractable \cite{YuLav15TRO-I}. The apparent similarity between optimal $\mpp$ and multiflow is perhaps best explained through a graph-based reduction from $\mpp$ problems to network flow problems. The reduction will also form the basis of our algorithmic solution. 

\begin{figure}[htp]
\begin{center}
  \begin{tabular}{cc}
    \includegraphics[height=0.20\textwidth]{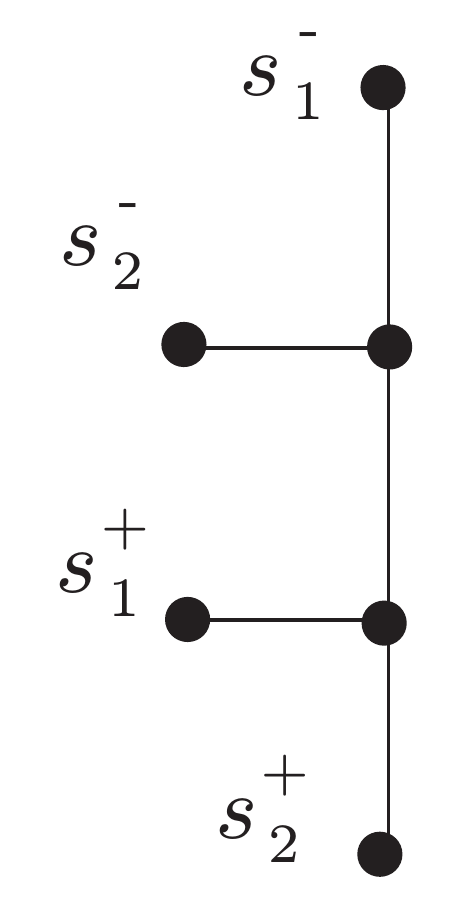} & 
    \includegraphics[height=0.20\textwidth]{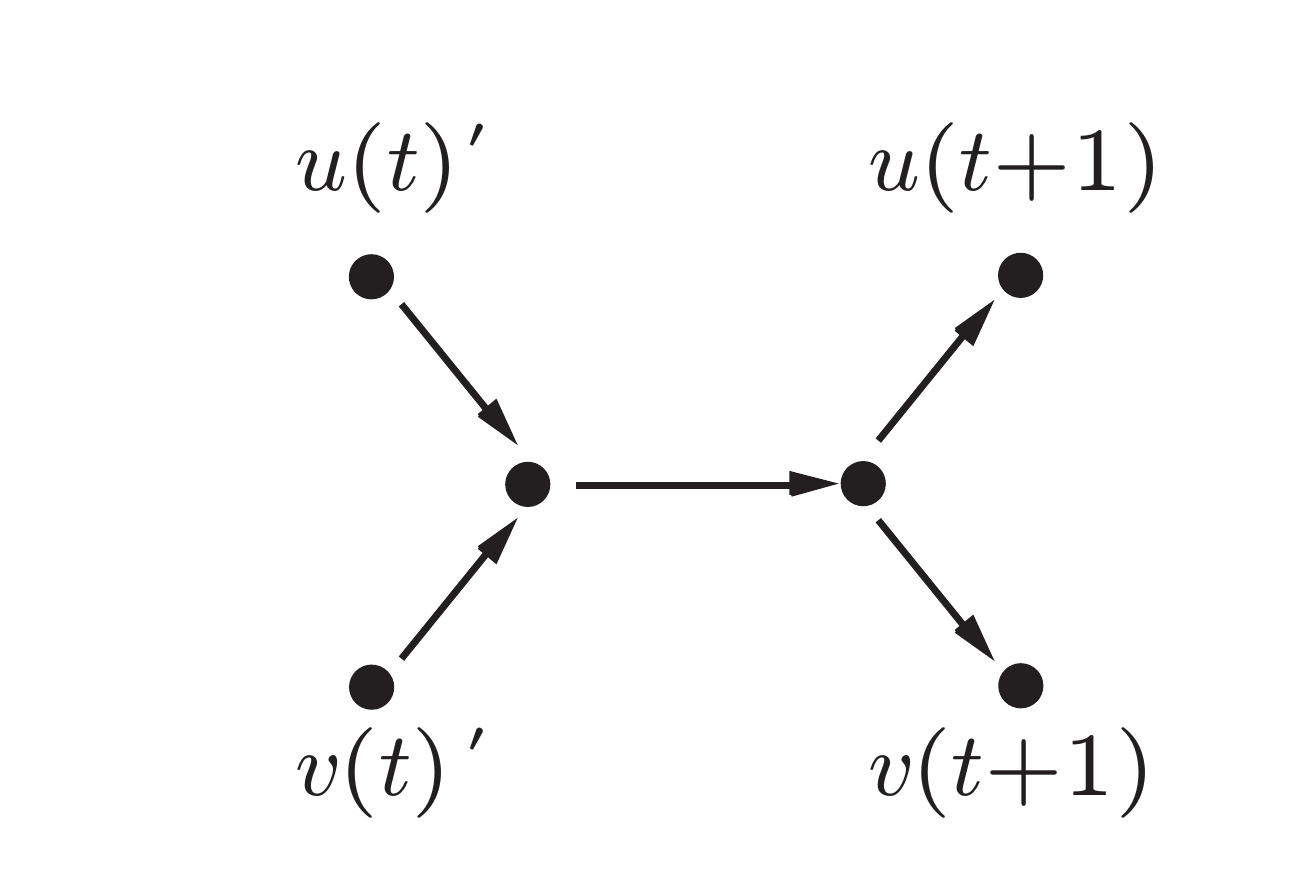}  \\
    (a) & (b)\\
  \end{tabular}
\end{center}
\caption{\label{fig:pimpp} a) A simple $G$. b) A {\em merge-split} gadget for splitting an undirected edge through time steps, for enforcing the head-on collision constraint.}
\end{figure}

To describe the reduction, we use as an example the undirected graph $G$ in Fig. \ref{fig:pimpp}(a), with start vertices $\{s_i^+\}, i = 1, 2$ and goal vertices $\{s_i^-\}, i = 1, 2$. An instance of $\mpp$ is given by $(G, \{r_1, r_2\}, x_I: r_i \mapsto s^+_i, x_G: r_i \mapsto s^-_i)$. We will reduce the problem to a network flow problem $\mathcal N = (G', c_1, c_2, S)$. 
The reduction proceeds by constructing a network that is a {\em time-expanded} version of the graph $G$, which then allows the explicit consideration of the interactions among the robots over space {\em and} time. To carry out this expansion, a time horizon must first be decided. For different optimality objectives, the expansion time horizon, some natural number $T$, is generally different; for now we assume that $T$ is fixed. 

\begin{figure}[htp]
\begin{center}
    \includegraphics[width=0.44\textwidth]{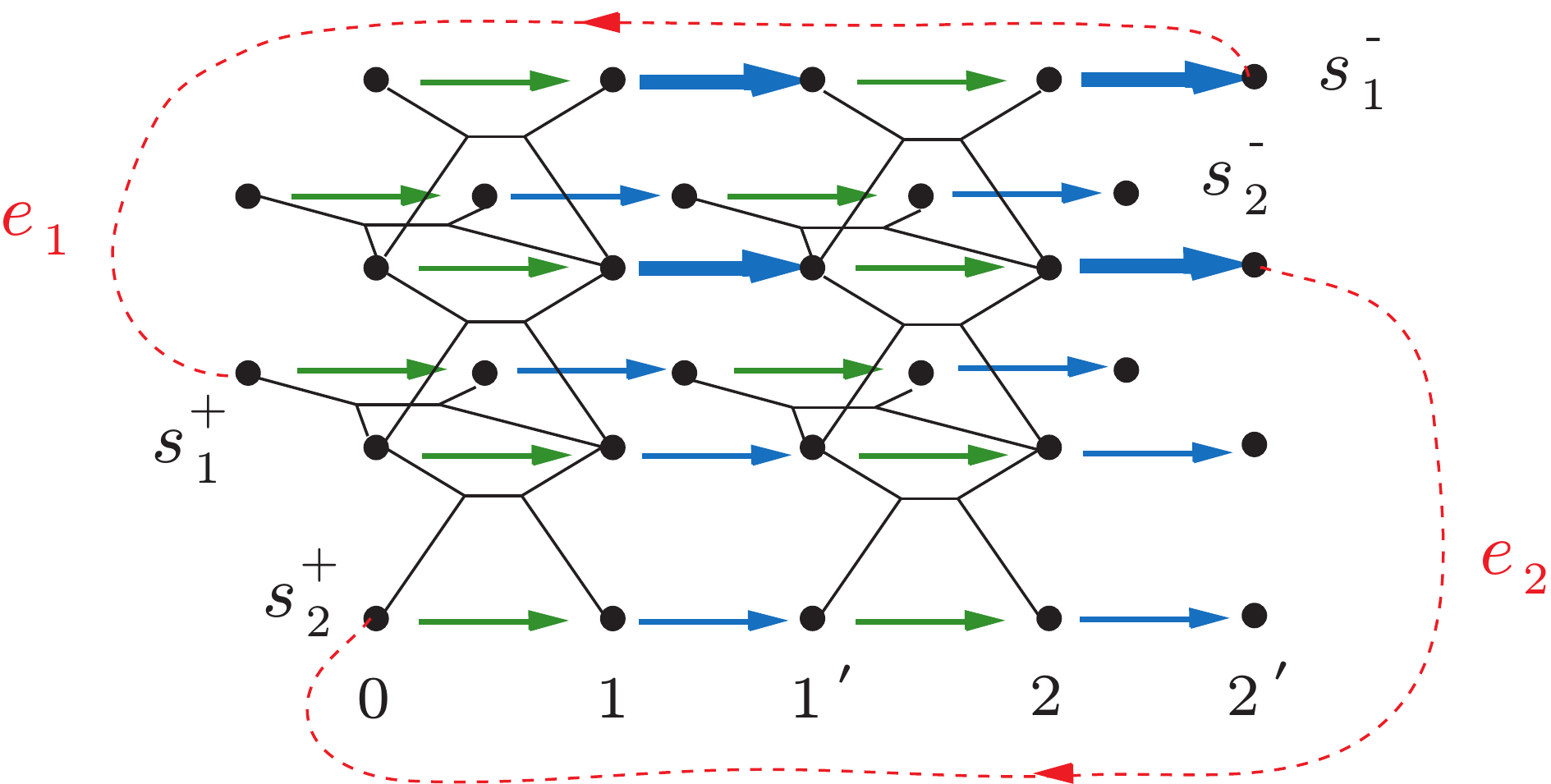}
\end{center}
\caption{\label{fig:pimpp-n} The time-expanded network with an expansion time horizon of $T = 2$  over the base graph Fig.~\ref{fig:pimpp}(a).}
\end{figure}

To begin building the network, we create $2T+1$ copies of $G$'s vertices, with indices $0, 1, 1', \ldots$, as shown in Fig. \ref{fig:pimpp-n}. For each vertex $v \in G$, denote these copies $v(0) = v(0)', v(1), v(1)', v(2), \ldots, v(T)'$. For each edge $(u, v) \in G$ and time steps $t, t+1$, $0 \le t < T$, the {\em merge-split} gadget shown in Fig. \ref{fig:pimpp}(b) is added between $u(t)', v(t)'$ and $u(t+1), v(t+1)$ (arrows from the gadget are omitted from Fig. \ref{fig:pimpp-n} since they are small). For the gadget, we assign unit capacity to all edges, unit cost to the horizontal middle edge, and zero cost to the other four edges. The merge-split gadget ensures that two robots cannot travel in opposite directions on an edge in the same time step, which prevents head-on collision between two robots. To finish the construction of Fig. \ref{fig:pimpp-n}, for each vertex $v \in G$, we add one edge between every two successive copies ({\em i.e.}, we add the edges $(v(0),v(1)), (v(1), v(1)'), \ldots, (v(T), v(T)')$). These correspond to the green and blue edges in Fig. \ref{fig:pimpp-n}. For all green edges, we assign them unit capacity and cost; for all blue edges, we assign them unit capacity and zero cost. The green edges allow robots to stay at a vertex during a time step whereas blue edges ensure that each vertex holds at most one robot, enforcing the meet collision constraint. 

Fig. \ref{fig:pimpp-n} (with the exception of edges $e_1$ and $e_2$, which will become relevant shortly), the time-expanded network, is the desired $G'$. For the set $S = S^+ \cup S^-$, we may simply let $S^+ = \{v(0): v \in \{s^+_i\} \}$ and $S^- = \{v(T)': v \in \{s^-_i\}\}$. $\mathcal N = (G', c_1, c_2, S)$ is now complete; we have reduced $\mpp$ to an integer multiflow problem on $\mathcal N$ with each robot from $R$ as a single type of commodity. 

\begin{theorem}\label{t:mpp}Let $(G, R, x_I, x_G)$ be an $\mpp$ instance. There is a bijection between its solution set (with a maximum number of time steps up to $T$) and the integer maximum multiflow solutions of flow value $n$ on the time-expanded network $\mathcal N$ constructed from $(G, R, x_I, x_G)$ with $T$ time steps. 
\end{theorem}
{\sc Proof.} {\bf Injectivity}. Assume that $P = \{p_1, \ldots, p_n\}$ ($n$ is the number of robots) is a solution to an instance of $\mpp$. For each $p_i$ and every time step $t = 0, \ldots, T$, we mark the copy of $p_i(t)$ and $p_i(t)'$ (recall that $p_i(t)$ corresponds to a vertex of $G$) at time step $t$ in the time-expanded graph $G'$. Connecting these vertices of $G'$ sequentially (there is a unique way to do this) yields one unit of flow $f_i$  on $\mathcal N$ (after connecting to appropriate source and sink vertices in $S^+, S^-$, which is trivial). It is straightforward to see that if two paths $p_i, p_{j}$ are not in collision, then the corresponding flows $f_i, f_j$ on $\mathcal N$ are vertex disjoint paths and therefore do not violate any flow constraint. Since any two paths in $P$ are not in collision, the corresponding set of flows $\{f_1, \ldots, f_n\}$ is feasible and maximal on $\mathcal N$. 

{\bf Surjectivity}. Assume that $\{f_1,\ldots, f_n\}$ is an integer maximum multiflow on the network $\mathcal N$ with $|f_i| =1$ for all $i$'s. First we establish that any pair of flows $f_i, f_j$ are vertex disjoint. To see this, we note that $f_i, f_j$ (both are unit flows) cannot share the same source or sink vertices due to the unit capacity structure of $\mathcal N$ enforced by the blue edges. If $f_i, f_j$ share some non-sink vertex $v$ at time step $t > 0$, both flows then must pass through the same blue edge (see Fig. \ref{fig:pimpp}(b)) with $v$ being either the head or tail vertex, which is not possible. Thus, $f_i, f_j$ are vertex disjoint on $\mathcal N$. We can readily convert each flow $f_i$ to a corresponding path $p_i$ (after deleting extra source vertex, sink vertices, vertices in the middle of the gadgets, and tail vertices of blue edges) with the guarantee that no $p_i, p_j$ will collide due to a meet collision. By construction of $\mathcal N$, the gadget we used ensures that a head-on collision is also impossible. The set $\{p_1, \ldots, p_n \}$ is then a solution to $\mpp$. ~\qed

\remark A multiflow problem can be reduced to an $\mpp$ problem (on a directed graph) as well. In particular, if all edges in the network have unit capacity and there is a single unit of each type of commodity, then a multiflow problem {\em is} a multi-robot path planning problem, often known as the {\em edge disjoint path} problem \cite{RobSey95}. ~\rqed

\section{Complete, Integer Linear Programming-Based Algorithms for Optimal $\mpp$ Problems}\label{sec:model}

Because optimizing $\mpp$ solutions over Objectives~\ref{omakespan}-\ref{otd} are computationally intractable, reducing $\mpp$ to multiflow problems does not make these optimal $\mpp$ problems any easier. However, with a network flow formulation (see Secion~\ref{subsection:network-flow-review}), it becomes possible to establish integer linear programming (ILP) models for optimal $\mpp$ formulations. These ILP models can then be solved with powerful linear programming packages. In comparison to A${}^*$-based algorithms augmented with heuristics which often target important but a limit set of problem structures, ILP-based algorithms proposed here are {\em agnostic} to specific problem structures. As such, ILP-based algorithms appear more capable at attacking a wider range of $\mpp$ problems and in particular  difficult $\mpp$ instances in which the robot-vertex ratio is high. In this section, we build ILP models for each of Objectives~\ref{omakespan}-\ref{otd}.

\subsection{Minimizing the Makespan}
A minimum makespan solution to an $\mpp$ instance $I = (G, R, x_I, x_G)$ can be computed using a {\em maximum multiflow} formulation. Fixing a time span $T$, let $\mathcal N = (G', c_1, c_2, S)$ be the time-expanded network for $I$, a set of $n$ {\em loopback} edges are added to $G'$ by connecting each pair of corresponding start and goal vertices in $S$, from the goal to the start. We use $e_j$'s to denote edges of $G'$ and let the $n$ loopback edges take the first $n$ indices, with $e_j, 1 \le j \le n$ being the edge connecting the goal vertex of $r_j$ to the start vertex of $r_j$. For example, for the $G'$ in Fig.~\ref{fig:pimpp-n}, $e_1$ and $e_2$ are the loopback edges for $r_1$ and $r_2$, respectively. The lookback edges have unit capacities and zero costs. Next. for each edge $e_j \in G'$ (including the lookback edges), create $n$ binary variables $x_{1, j}, \ldots, x_{n,j}$ corresponding to the flow through that edge, one for each robot. The is, $x_{i, j} = 1$ if and only if robot $r_i$ passes through $e_j$ in $G'$. The variables $x_{i,j}$'s must satisfy two edge capacity constraints and one flow conservation constraint, 
\begin{equation}\label{to1}
\begin{array}{cc}
\forall\, e_j, & \displaystyle\sum_{i=1}^n x_{i,j} \le 1,\\
\forall\, 1 \le i, j \le n, \,i \ne j, & \displaystyle x_{i, j} = 0, 
\end{array}
\end{equation}
\begin{equation}\label{to2}
\forall\, v \in G' \textrm{ and } 1 \le i \le n, \displaystyle\sum_{e_j \in \delta^+(v)} x_{i,j} = \sum_{e_j \in \delta^-(v)} x_{i,j}.
\end{equation}
The objective function is 
\begin{equation}\label{to3}
\max \sum_{1 \le i \le n} x_{i,i}.
\end{equation}

For each fixed $T$, a solution to the above ILP model with an objective equaling $n$ means that a feasible solution to $\mpp$ is found. We are to find the minimal $T$ that yields such a feasible solution. To do this, start with $T$ being the maximum over all robots the shortest possible path length for each robot, ignoring all other robots. An ILP model for this $T$ is then built and tested for a feasible solution. If the model is not feasible, $T$ is increased and the procedure repeated. The first feasible $T$ is the optimal $T$. Once a feasible model with the minimum expansion time horizon $T$ is found, the robots' paths can be extracted based on the proof of Theorem \ref{t:mpp}. The algorithm is in fact complete: Since the problem is discrete, there is only a finite number of possible states. Therefore, for some sufficiently large $T$, there must either be a feasible solution or we can pronounce that none can exist ($T = O(|V|^3)$ is big enough \cite{YuRus14WAFR}). Denoting the resulting algorithm as \tompp\, we have shown the following.
\begin{proposition}\label{p:time}Algorithm \tompp\, is a complete algorithm for finding minimum makespan solutions for $\mpp$.
\end{proposition}

%\textbf{ADD AN ALGORITHM FLOW CHART?}
\subsection{Minimizing the Maximum Single-Robot Traveled Distance}
For minimizing the maximum distance traveled by any single robot, the network and variable creation remains the same as the minimum makespan setup; constraints~\eqref{to1} and~\eqref{to2} remain unchanged. Because we want to send all robots to their goal, a maximum flow may be forced through the additional constraint
\begin{equation}\label{c:flow}
\forall 1 \le i \le n,\quad x_{i,i} = 1.
\end{equation}

To encode the min-max objective function, we introduce an additional integer variable $x_{max}$ and add the constraint 
\begin{equation}\label{c:xmax}
\sum_{e_j \in G', j > n} c_2(e_j) \cdot x_{i,j} \le x_{max}
\end{equation}
for all $1 \le i \le n$. For a fixed $i$, the left side of~\eqref{c:xmax} represents the distance traveled by robot $r_i$. The objective function is then simply 
\begin{equation}\label{md-o4}
\min x_{max}. 
\end{equation}

Denoting the algorithm as \mdmpp, we have
\begin{proposition}Algorithm \mdmpp\, is a complete algorithm for finding solution to $\mpp$ that minimize the maximum distance traveled by a single robot.
\end{proposition}
\remark Assuming that we have a minimum makespan solution, we can better bound the time horizon $T$ needed for \mdmpp. Let $t_{min}$ be the minimum makespan computed by \tompp, setting $T = nt_{min}$ is then sufficient for finding the $\min x_{max}$. To see that this is true, we first note that $\min x_{max} \le t_{min}$ because the minimum makespan solution requires robots to to synchronize their moves, which may force some robots to travel unnecessarily. Then, the $n$ robots cannot move a total distance of more than $n\min x_{max} \le nt_{min}$ because no robot may travel more than $\min x_{max}$ edges and therefore cannot use more than $n\min x_{max}$ time in total. ~\rqed

\subsection{Minimizing the Total Arrival Time}
The ILP model for minimizing the total arrival time is more involved in the way the objective function is constructed. First, the network and all variables from the minimum makespan ILP-model are inherited. We also inherit constraints~\eqref{to1},~\eqref{to2}, and~\eqref{c:flow}. To represent the objective function, for each time step $1 \le t \le T$ and each $v := x_G(r_i)$, $1 \le i \le n$, we create a binary variable $y_i^t$. Then, we give new indicies to some variables that are already created. Recall that for each edge $e_j = (v(t), v(t)') \in G'$ ({\em e.g.}, the four extra bold blue edges in Fig.~\ref{fig:pimpp-n}), a variable $x_{i,j}$ is created. There are a total of $nT$ such variables. Here, we give these variable a second index $x_i^t$. That is, $x_i^t$ is the binary variable indicating whether edge $(v(t), v(t)') \in G'$, $v := x_G(r_i)$, is used by robot $r_i$. 

Given a network with a fixed $T$, if constraints~\eqref{to1},~\eqref{to2}, and~\eqref{c:flow} can be satisfied, then there is a feasible solution to the original $\mpp$ problem. In this case, $x_i^T \equiv 1$ for $1 \le i \le n$. We let $y_i^T = x_i^T$. Then, each $y_i^t$, $1 \le t < T$ is defined recursively over $x_i^t$ and $y_i^{t+1}$ as 
\begin{equation}\label{c:and}
y_i^t \ge y_i^{t+1} + x_i^t - 1,\quad y_i^t \le y_i^{t+1},\quad y_i^t \le x_i^t.
\end{equation}

The constraint~\eqref{c:and} effectively performs the logical {\em and} operation over $x_i^t$ and $y_i^{t+1}$ and stores the result in $y_i^t$. In the end, the smallest $t$ for which $y_i^t = 1$ is the time robot $r_i$ reaches its goal (and stops). Therefore, for each $1 \le i \le n$, $\sum_{t = 1}^Ty_i^t$ is the number of time steps from the time $r_i$ arrives at its goal until time $T$. $T - \sum_{t = 1}^Ty_i^t$ is then the time spent by $r_i$. To minimize the total arrival time, the objective function can be expressed as 
\begin{equation}\label{of:tt}
\displaystyle nT - \sum_{1 \le i \le n, 1 \le t \le T}y_i^t.
\end{equation}

Denoting the resulting algorithm (with a sufficiently large $T$) as \ttmpp, we have
\begin{proposition}Algorithm \ttmpp\, is a complete algorithm for finding minimum total arrival time solutions for $\mpp$.
\end{proposition}

\subsection{Minimizing the Total Distance}
From the ILP model for minimizing the maximum distance, we only need to change the objective function for computing a minimum total distance solution. We do not need the variable $x_{max}$ and simply update the objective function to
\begin{equation}\label{to4}
\min \sum_{e_j \in G', j > n,\, 1 \le i \le n} c_2(e_j) \cdot x_{i,j}.
\end{equation}

Denoting the algorithm as \dompp, we have 
\begin{proposition}Algorithm \dompp\, is a complete algorithm for finding minimum total distance solutions for $\mpp$.
\end{proposition}
Again, $T = nt_{min}$ is sufficient for building a network that contains a minimum total distance solution, if one exists. 

\section{Heuristics for Effective Computation of Near-Optimal Solutions}\label{section:heuristics}
In Section~\ref{sec:model}, in constructing the ILP models for Objectives~\ref{omakespan}-\ref{otd}, our goal is to show the universal applicability of the network flow model ({\em e.g.}, Fig.~\ref{fig:pimpp-n}) toward many different optimal $\mpp$ formulations. The resulting ILP-based algorithms are complete and always produce the optimal solution in principle. As will be shown in Section~\ref{sec:evaluation}, such algorithms perform very well in handling relatively small but extremely challenging problems. Nevertheless, as the problem size grows ({\em i.e.}, as the graph $G$ and the number of robots $n$ get larger), the computation time needed by ILP solvers grows rapidly. From a practical point of view, it may be far more desirable to quickly compute a good quality but sub-optimal solution than waiting forever for the optimal solution. In this section, we introduce several heuristics to accomplish just that, with a particular focus on computing solutions with minimum makespan. 

\subsection{Building More Compact Models}
To get the best performance out of a solver, it is beneficial to have a lean model ({\em i.e.}, fewer columns and rows). So far, our focus has been to provide a general network flow based framework so that the ILP models can be easily built. When it comes to translating the models to an ILP solver, they can be further simplified. The heuristics discussed in this subsection aim at making the representation of the constraints~\eqref{to1} and~\eqref{to2} more compact in the resulting ILP models. As such, they apply to all the optimality objectives.

\subsubsection*{Better encoding of the collision constraints}
\begin{figure}[htp]
\begin{center}
    \includegraphics[width=0.44\textwidth]{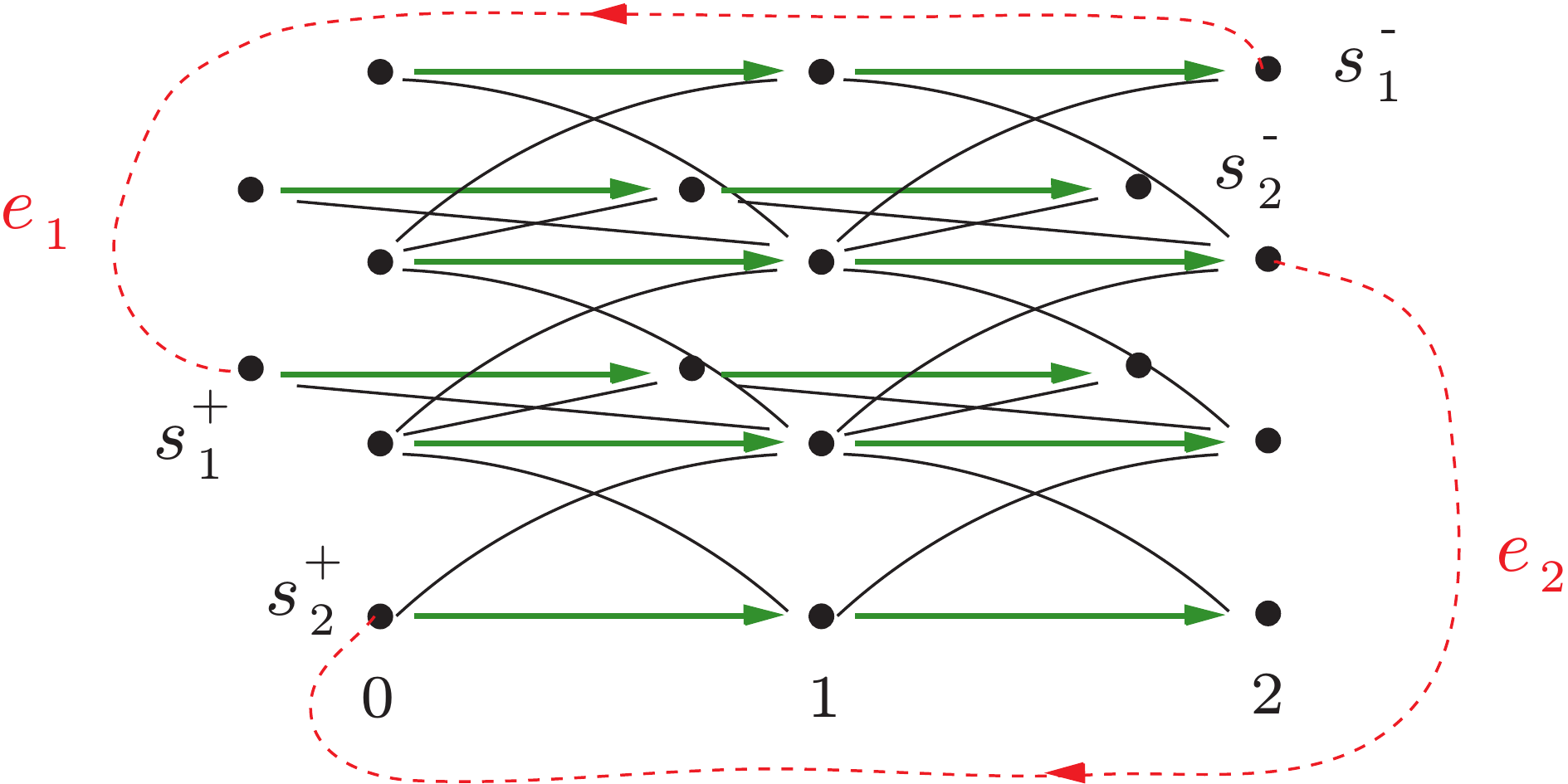}
\end{center}
\caption{\label{fig:pimpp-n-compact} A more compact representation of the network flow graph from Fig.~\ref{fig:pimpp-n}.}
\end{figure}

Recall that in building the network flow model ({\em e.g.}, Fig.~\ref{fig:pimpp-n}), we used a merge-split gadget (Fig.~\ref{fig:pimpp}(b)) for enforcing the head-on collision constraint and extra time steps ({\em e.g.}, the blue edges in Fig. \ref{fig:pimpp-n}) for avoiding meet collisions. When we translate this into linear constraints, these structures can be simplified to yield the more compact structure illustrated in Fig.~\ref{fig:pimpp-n-compact}. 

\begin{figure}[htp]
\begin{center}
    \includegraphics[height=0.20\textwidth]{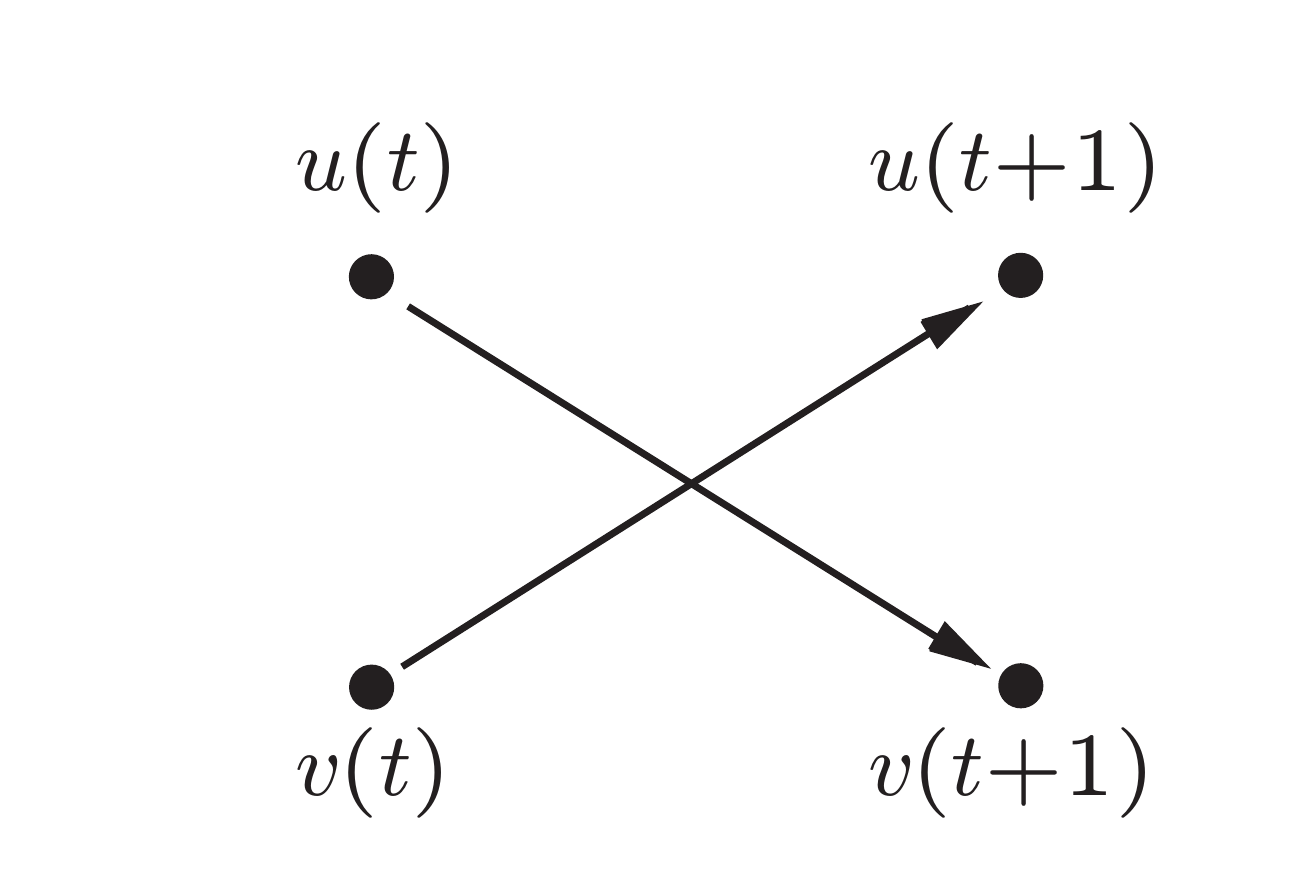}  
\end{center}
\caption{\label{fig:pimpp-b} The simplified {\em merge-split} gadget for enforcing the head-on collision constraint.}
\end{figure}

In the newer structure, each merge-split gadget now has two edges instead of five. Also, the blue edges are removed. The updated gadget for an edge $(u, v) \in E$ between time steps $t$ and $t+1$ is shown in Fig.~\ref{fig:pimpp-b} (note that due to the removal of the blue edges, vertices such as $v(t)'$ is no longer needed). 
That is, instead of five, only two variables are needed for each robot. Denoting these binary variables as $x_{i, (u(t), v(t+1))}$ and $x_{i, (v(t), u(t+1))}$ for a robot $r_i$, the head-on collision constraint for a single gadget can be readily encoded as 
\begin{equation}\label{c:head-on}
\displaystyle\sum_{i = 1}^n x_{i, (u(t), v(t+1))} + \sum_{i = 1}^n x_{i, (v(t), u(t+1))} \le 1.
\end{equation}

Then, to enforce the meet constraint, for example at a vertex $v(t)$, we simply require that at most one outgoing edge from $v(t)$ may be used, {\em i.e.},
\begin{equation}\label{c:meet}
\displaystyle\sum_{e_j \in \delta^+(v(t)), 1 \le i \le n} x_{i, j} \le 1.
\end{equation}

Overall, the newer ILP model is roughly half of the size of the original model. 
\subsubsection*{Reachability analysis} In the time-expanded graph, there are redundant binary (edge) variables that can never be true because some edges are never reachable. For example, in Fig.~\ref{fig:pimpp-n-compact}, at $t = 0$, the only outgoing edges that can possible be used are those originates from $s_1^+$ and $s_2^+$. The rest can be safely removed. In general, for each robot $r_i$, based on its reachability from its start vertex and to its goal vertex, a sizable number of binary variables $x_{i,j}$'s can be deleted. 

\subsection{Divide-and-Conquer Over Time Domain}\label{subsection:splitting}
\begin{figure}[htp]
\begin{center}
  \begin{tabular}{ccc}
    \includegraphics[width=1.05in]{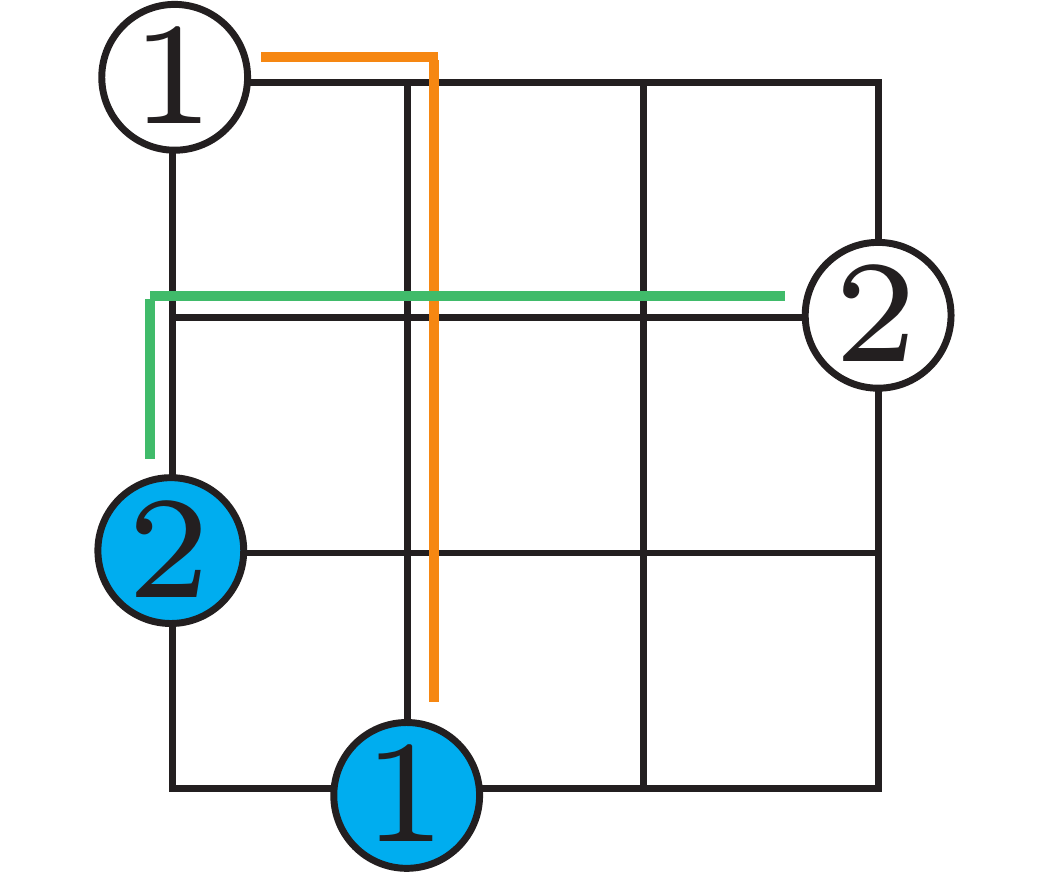} & \hspace{10mm} &
		\includegraphics[width=1.05in]{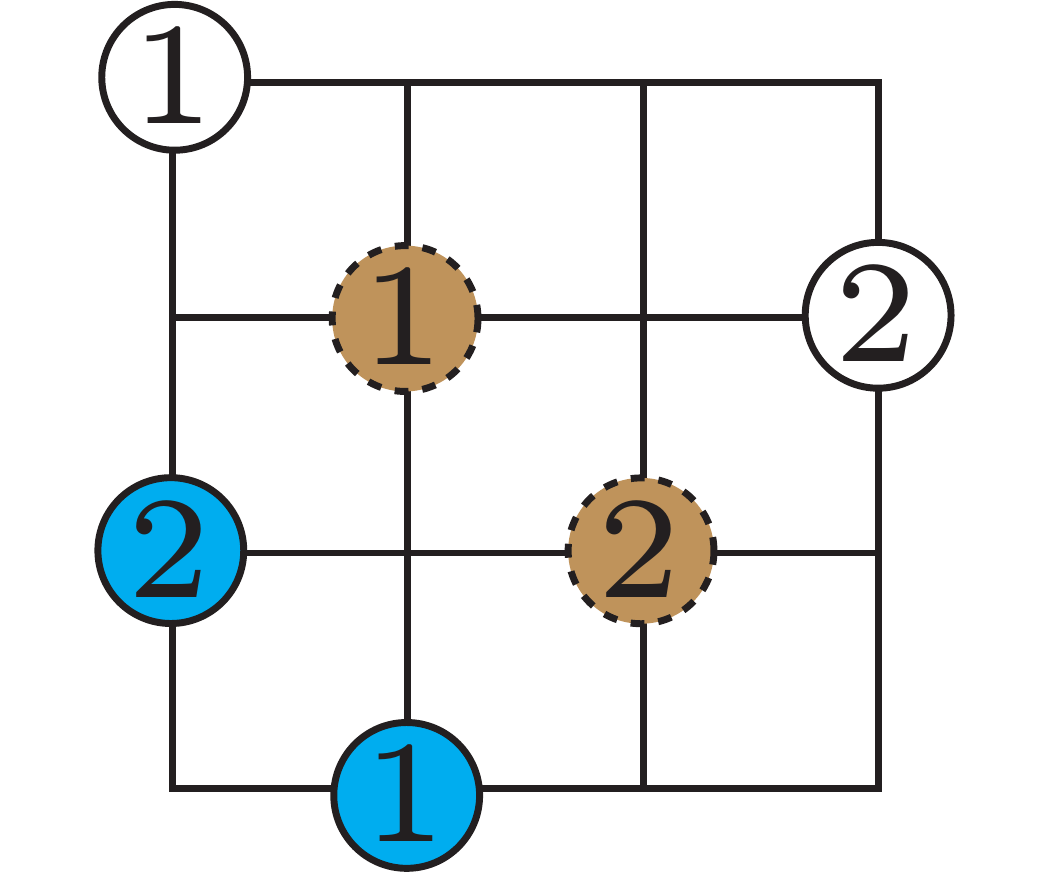}\\
		(a) & & (b)\\
  \end{tabular}
\end{center}
\vspace*{-2mm}
\caption{a) A simple two-robot problem. b) The time-divided instances.} \label{fig:dac}
\end{figure}

In evaluating the ILP model-based algorithm for optimal makespan computation, we observe that the ILP solver running time appears to grow exponentially as the size of the model grows. This prevents the algorithm from performing well over instances with more than a few tens of robots. The observation, while hampering the effectiveness of the exact algorithm, turns out to offer a useful insight toward a highly effective heuristic. We find that, when the overall size of the ILP model is relatively small and the robot-vertex ratio is not approaching $1$, even when there are a large number of robots, the model usually does not present much challenge for ILP solvers. To apply the ILP model-based method to more challenging problems ({\em e.g.}, solving problems with hundreds of robots quickly), we simply limit the size of the individual ILP model fed to the solver. One way to achieve this is through {\em divide-and-conquer over the time domain}. We use a simple example (see Fig.~\ref{fig:dac}) to illustrate the idea. 

In Fig.~\ref{fig:dac}(a), we have a simple planning problem for two robots on a $3 \times 3$ grid. To execute the heuristic, we first compute a shortest path for every pair of start and goal locations. In this case, we get the orange and green paths for robots $1$ and $2$, respectively. Then, if we decide to split the problem into two smaller problems, for each of the paths, it is split into two (generally) equal length pieces and the middle node is set as the intermediate goal. In our example, we may do this for robot $1$ by setting the intermediate goal location at $(1, 1)$ from the top-left corner (the brown disc labeled $1$ in Fig.~\ref{fig:dac}(b)). For robot $2$, because the middle location coincides with that of robot $1$, we pick an alternative location that is not already occupied as the intermediate goal for robot $2$, in this case $(2, 2)$ from the top-left corner. The intermediate goals for the first instance will also serve as the start locations of the second instance. This yields two child instances with both requiring a time expansion with $2$ steps each, effectively making the individual ILP model roughly half the size of the original one, which requires a time expansion with $4$ steps. In general, we may divide a problem into arbitrarily many smaller instances in the time domain. 

If a problem is divided in this manner to $k$ sub problems, we call the resulting heuristic a {\em $k$-way split}. Because the division is over time, there is no interaction between the individual, smaller instances. Once we obtain the solution for each child instance, the solutions can be glued together by concatenating the results. In practice, this simple heuristic dramatically improves algorithm performance without heavy negative impact on path optimality in terms of makespan; we observe a consistent speedup in computational experiments. \,\,\,\,\,

\vspace{1mm}
\remark The $k$-way split heuristic, by design, is particularly suitable for the makespan objective. This is the case due to the additive nature of the makespan objective over the split sub-problems. Besides makespan, the heuristic also applies to distance objectives ({\em i.e.}, Objectives~\ref{md} and~\ref{otd}) quite well, as long as the time horizon required for finding distance optimal solution does not differ greatly from the time horizon required for minimum makespan solution. The heuristic does not directly apply to Objective~\ref{ott} because total time is not additive over the split sub-problems. As an example, suppose that a 2-way split is carried out with each sub-problem having a time horizon of $T/2$. If a robot $r_i$ does not move in the solution to the first sub-problem ({\em i.e.}, $0 \le t \le T/2$), it contributes $0$ to the total distance. However, if $r_i$ moves even a single step in the solution to the second sub-problem ({\em i.e.}, $T/2 \le t \le T$), then $r_i$ will contribute at least $T/2$ to the total arrival time. Nevertheless, $k$-way split is still helpful in this case as we may use it to quickly compute an initial $T$ for performing the time expansion.~\rqed

\section{Experimental Evaluation}\label{sec:evaluation}
In this section, we evaluate the performance of our optimal and near-optimal $\mpp$ algorithms with an emphasis on \tompp. 
%given the particular importance of optimal makespan solutions. In the current context, minimizing makespan is equivalent to minimizing the task completion time, which is arguably a most important objective. Moreover, minimum makespan solutions are the gateways to other optimal solutions. As stated in Section~\ref{subsection:splitting}, if the time horizon is too long in the time-expanded network, it could bring a heavy hit to the performance of the ILP solver package. On the other hand, as long as a problem instance is not extremely challenging ({\em i.e.}, one with many robots on a poorly connected underlying graph), using the minimum makespan as the time horizon for building the time-expanded network proves to be effective in computing minimum total distance or minimum total arrival time solutions. Moreover, a minimum makespan solution is frequently also an optimal or near-optimal solution for minimizing the maximum single-robot travel distance. 
Our performance evaluation covers a broad spectrum of typical problem settings. For each setting, we push the limit on the robot-vertex ratio--to as high as $100\%$. To the best of our knowledge, the majority of the settings with high robot-vertex ratio have never been attempted with much success prior to our study. 

When applicable, we also compare our results with the state-of-the-art found in the literature. In particular, we have examined \odid, \ida, \wcha, and \cobopt\,\cite{Sur12}, among others. \odid\, and \ida\, support cycles whereas \wcha appears to be designed for cycle-free $\mpp$ as it could not solve any $9$-puzzle. The problem definition for \cobopt\, suggests it solves $\mpp$ but it employs a cycle-free subroutine for finding feasible solutions. \odid, \ida, and \wcha\, are designed for optimizing total time and total distance optimal objectives, and do not naturally extend to makespan computation. However, the associated makespan produced by these algorithms are usually of good quality. Among these three, our experiments show that \ida-based anytime algorithm is the most versatile due to its IDA*-like incremental structure. On the other hand, \odid\, and \wcha\, do not scale well when the robot-vertex ratio goes beyond $10$-$20\%$, depending on the particular problem setting.\footnote{Some of these algorithms were evaluated without requiring all robots reach their goals. For example, in the \wcha\, work \cite{Sil05}, if an instance with $n$ robots is solved for $p < n$ robots, the problem is counted as partially solved. We require each instance to be fully solved to be counted as a success.} \cobopt\, is designed for makespan computation. 

We implemented all algorithms (\tompp, \mdmpp, \ttmpp, and \dompp) in the Java programming language. We take advantage of multi-core CPUs when the $k$-way split heuristic is being used. Also, Gurobi \cite{gurobi}, the ILP solver used in our implementation, can engage multiple cores automatically for hard problems. We ran all our tests on a MacBook Pro laptop computer (Intel Core i7-4850HQ, 16GB memory). We thank Trevor Standley for sharing the C code implementing \odid, \ida, and \wcha, among others. We modified (the original code supports only $32\times 32$ grid) and compiled the code as a 64-bit windows executable under MSVC 2010 with all speed optimization flags turned on. The comparison to \cobopt\, uses the result provided in \cite{Sur12}, which covers only $8\times 8$ and $16 \times 16$ grids. 

\subsection{Performance of \tompp\,  and $k$-way Split Heuristic}\label{subsection:k-way-split}
\begin{figure}[htp]
\begin{center}
    \includegraphics[width=3.4in]{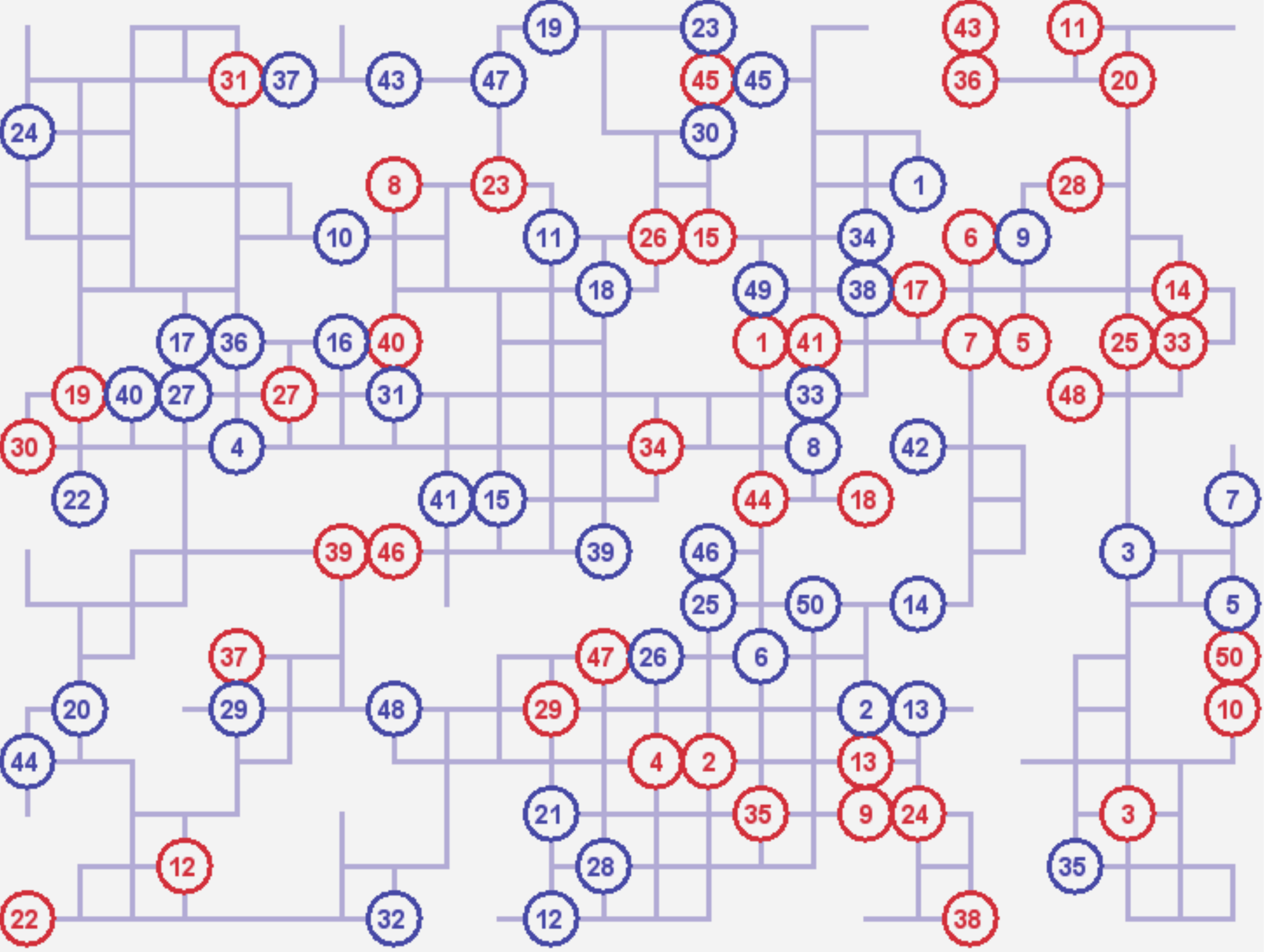} 
\end{center}
\vspace*{-2mm}
\caption{A typical $24 \times 18$ grid instance with $25\%$ vertices removed to simulate obstacles and 50 start and goal locations. Note that the connectivity of the graph is low at some areas. For example, the lower right corner blob is only singly-connected to the rest of the graph. Some (blue) start locations may overlap with some (red) goal locations.} \label{fig:24x18-30}
\end{figure}

We begin our experimental evaluation focusing on the \tompp\, algorithm and the $k$-way split heuristic. For this purpose, we use as the based graph a $24 \times 18$ grid with varying number of vertices ($0$-$25\%$) removed to simulate obstacles. The connectivity of the graph is always maintained. We note that with $25\%$ vertices removed, the graph is already sparsely connected at some places (see {\em e.g.}, Fig.~\ref{fig:24x18-30}), 
making solving these problems optimally a challenging task. In presenting the computational results, each data point in the figures is an average over $10$ sequentially, randomly generated instances. For each obstacle percentage, we start from the lowest number of robots (usually $10$ or $20$) and allocate a maximum of $600$ seconds for each problem instance. If an instance takes more than $600$ seconds to produce a result, the instance is failed and we move to the next obstacle percentage. This also means that a data point is given only if each of the $10$ instances is completed within $600$ seconds. Over the same set of problem instances, the \tompp\, algorithm is executed in the exact manner (which produces optimal makespan solutions) and with the $k$-way split heuristic.

\begin{figure}[htp]
\begin{center}
  \begin{tabular}{c}
    \includegraphics[width=2.8in]{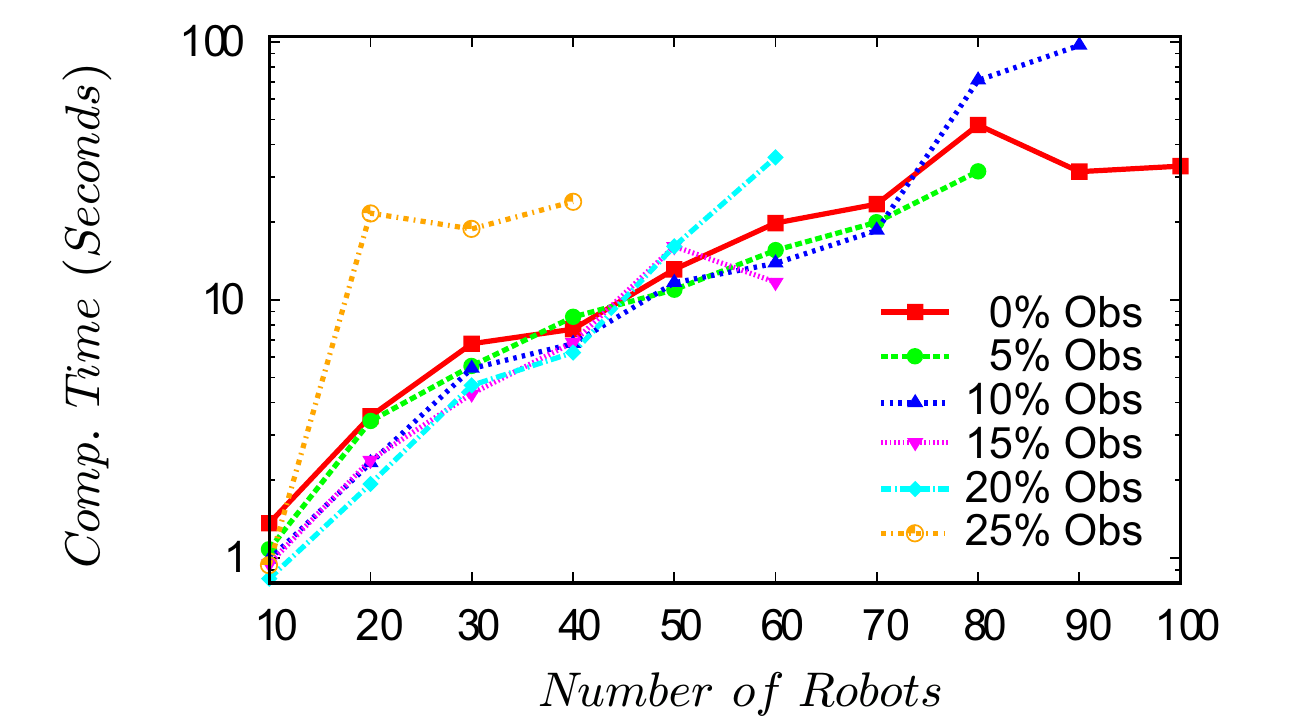} \vspace{1mm}\\
		\includegraphics[width=2.8in]{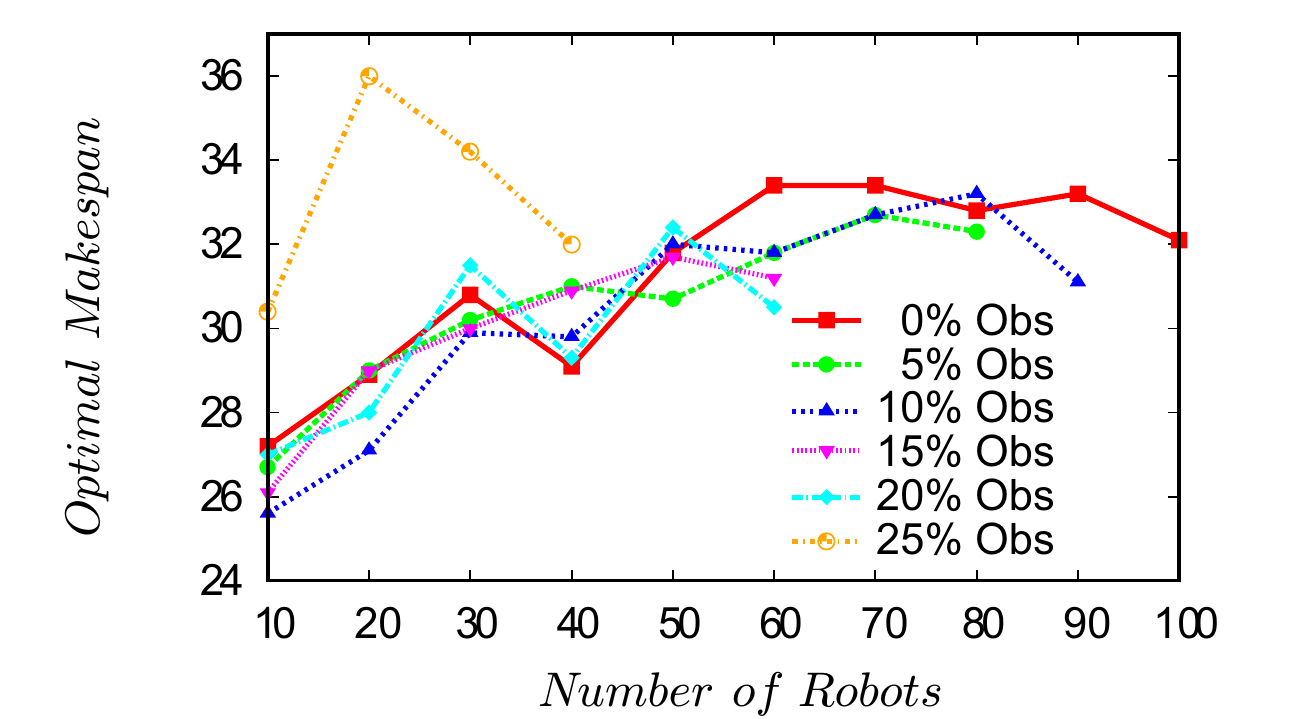}
  \end{tabular}
\end{center}
\vspace*{-2mm}
\caption{[top] Average computation time of the exact \tompp\, algorithm over instances on a $24 \times 18$ grid with randomly placed obstacles and start/goal locations. [bottom] The (average) optimal makespan.} \label{fig:24x18-ms}
\end{figure}

The exact makespan computation result is summarized in Fig.~\ref{fig:24x18-ms}. For all obstacle settings, the \tompp\, algorithm computes optimal makespan solutions consistently for up to $100$ robots with an average computation time of no more than $100$ seconds. notably, for $50$ robots and obstacles up to $20\%$, the \tompp\, algorithm is able to complete in about $15$ seconds in all cases. From the top plot of Fig.~\ref{fig:24x18-ms}, we observe that for each fixed obstacle percentage, the computation time appears to grow exponentially with respect to the number of robots. The computational difficulty of a particular problem instance also heavily depends on the actual optimal makespan. For example, a problem instance in the case of $25\%$ and $20$ robots has a particularly long makespan (see the bottom plot of Fig.~\ref{fig:24x18-ms}), resulting an unexpected jump of the computation time. 

\begin{figure}[htp]
\begin{center}
  \begin{tabular}{c}
    \includegraphics[width=2.8in]{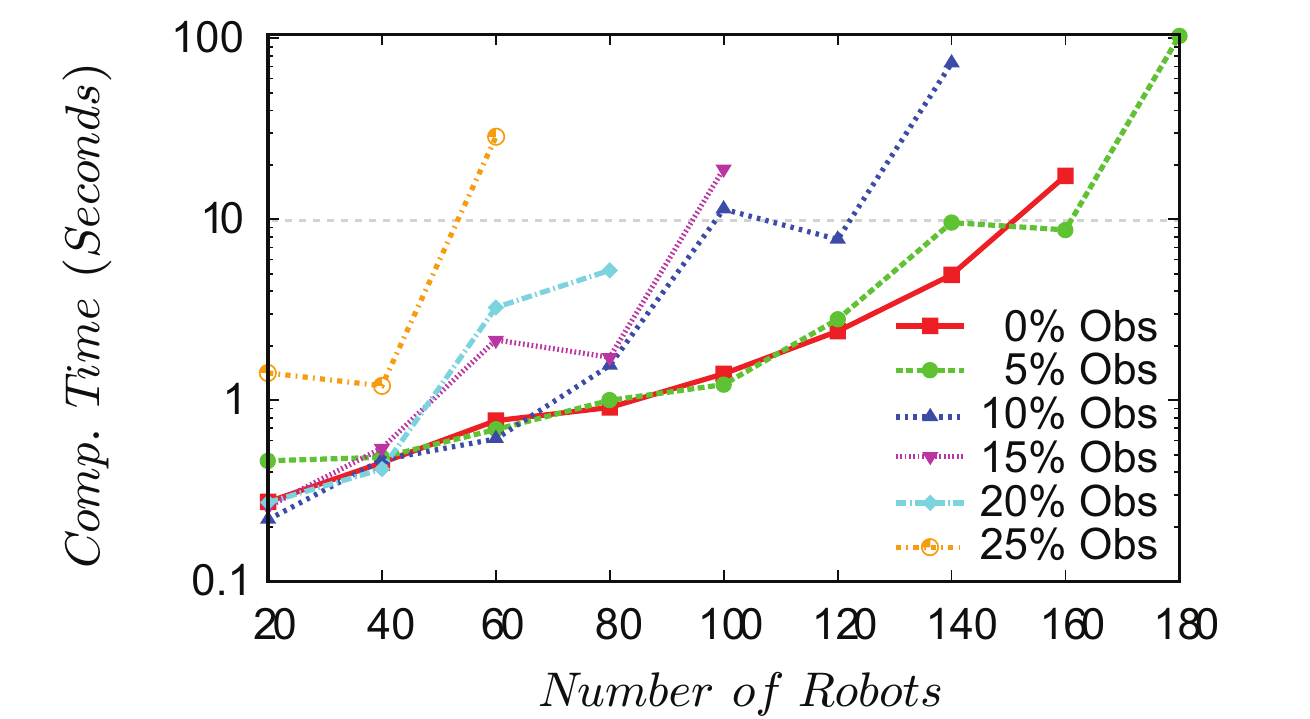} \vspace{1mm}\\
		\includegraphics[width=2.8in]{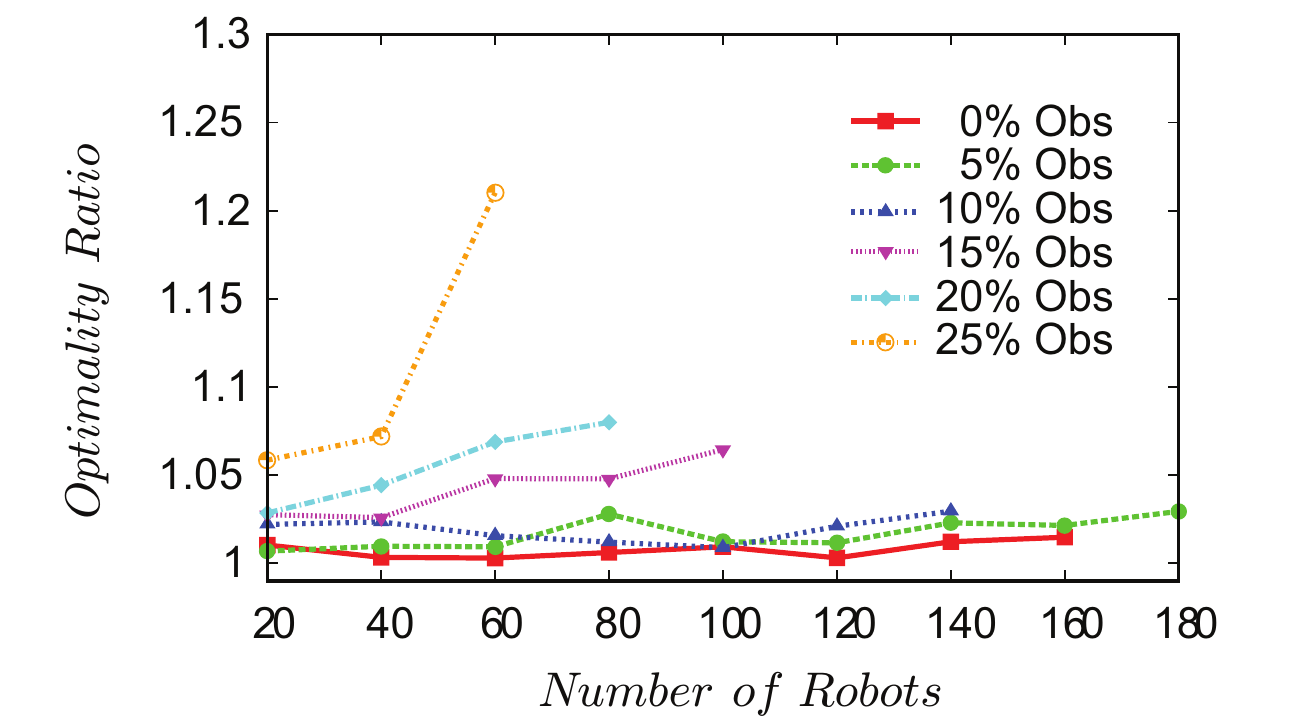}
  \end{tabular}
\end{center}
\vspace*{-2mm}
\caption{[top] Average computation time of \tompp\, algorithm (with $4$-way split heuristic) over instances on a $24 \times 18$ grid with randomly placed obstacles and start/goal locations. [bottom] The achieved (conservatively estimated) optimality ratio. } \label{fig:24x18-ms-4}
\end{figure}

Whereas the exact \tompp\, algorithm is reasonably efficient, the $k$-way split heuristic brings a significant performance boost, allowing a much higher robot-vertex density in general. In our tests, we are often able to double or triple the supported robot-vertex density. For the $24 \times 18$ grid, we evaluated the $k$-way split heuristic for $k$ up to $16$. The $4$-way split performance is illustrated in Fig.~\ref{fig:24x18-ms-4}. 
In the figure, we measure optimality using a conservatively estimated {\em optimality ratio}. To obtain this, we divide the objective value returned by the optimizer over a conservative estimate (a lower bound). For makespan, this lower bound estimate is obtained by first computing the shortest path for each robot ignoring all other robots. The minimum makespan estimate is obtained by taking the maximum length over all these shortest paths. Clearly, the optimality ratio obtained this way is an underestimate. 

We make three comments over Fig.~\ref{fig:24x18-ms-4}. First, from the top plot, we observe that our method is highly effective in terms of computation time, capable of computing minimum makespan solutions for up to $180$ robots, which translates to a maximum robot-vertex ratio of $44\%$. The majority of the cases are solved in under $10$ seconds. Even when there are $25\%$ obstacles, we could solve the problem consistently for up to $60$ robots in about $40$ seconds. Second, we again observe an exponential relationship between computation time and the number of robots. Third, all computed solutions are very close to being optimal, with all but one case having an optimality ratio of below $1.1$. The average minimum makespan for these instances is about $35$. 

\begin{figure}[htp]
\begin{center}
  \begin{tabular}{cc}
    \includegraphics[width=1.5in]{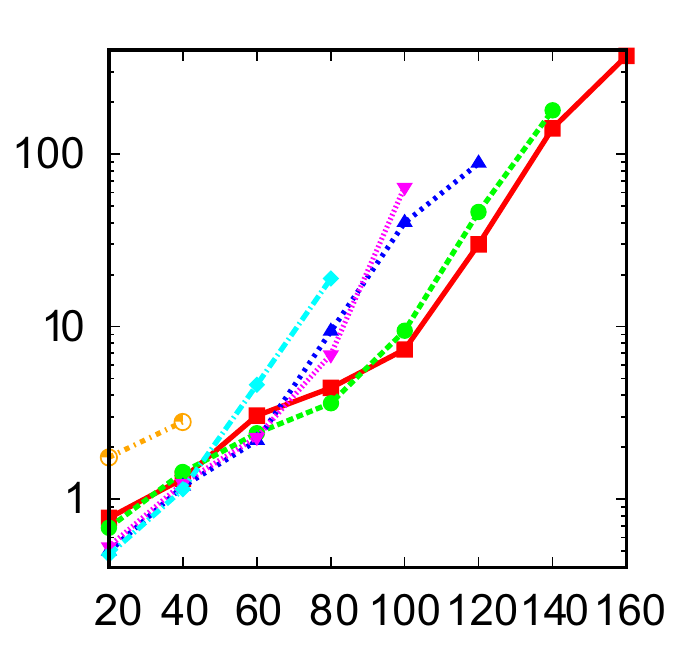} & 
		\includegraphics[width=1.5in]{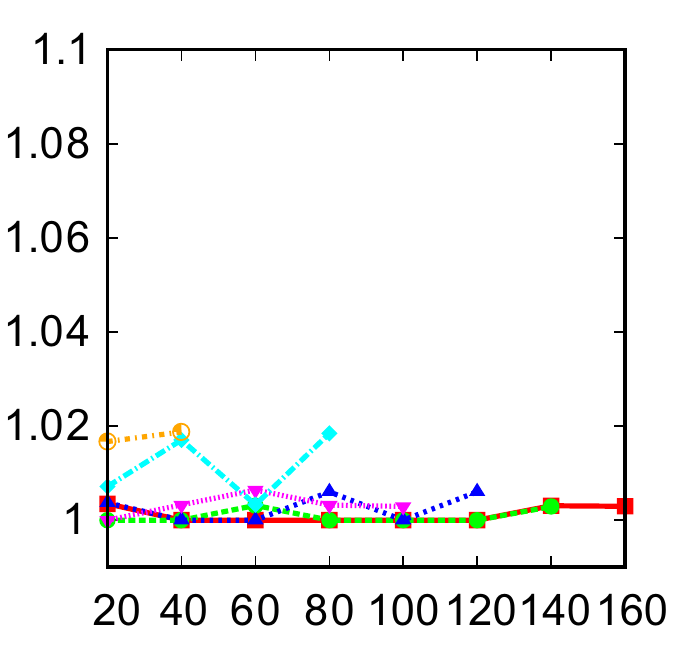} \vspace{1mm}\\
    \includegraphics[width=1.5in]{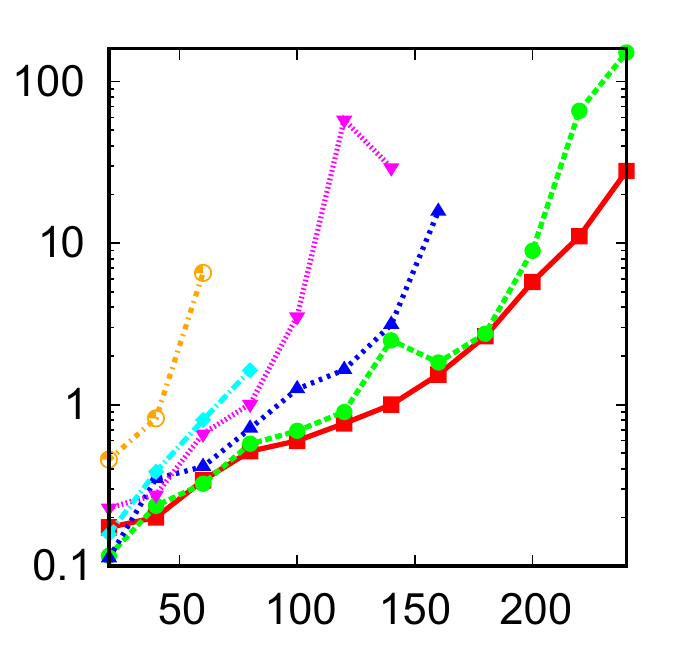} & 
		\includegraphics[width=1.5in]{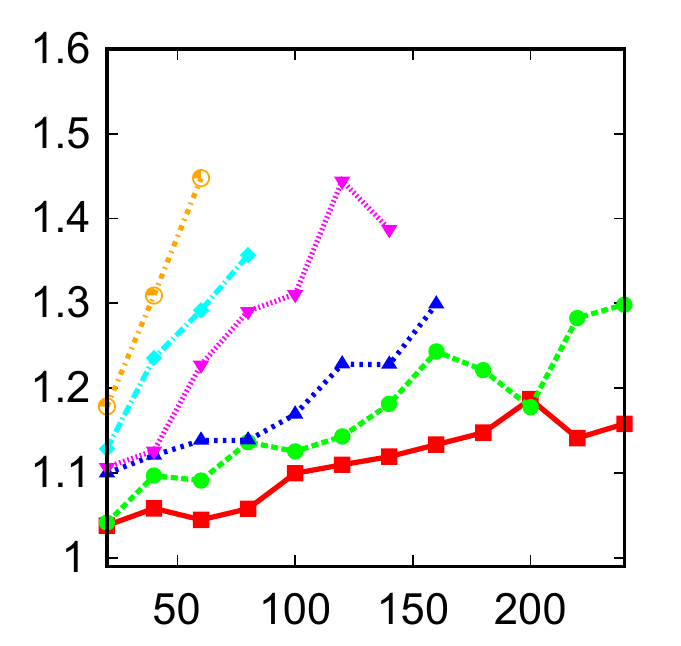} \vspace{1mm}\\
    \includegraphics[width=1.5in]{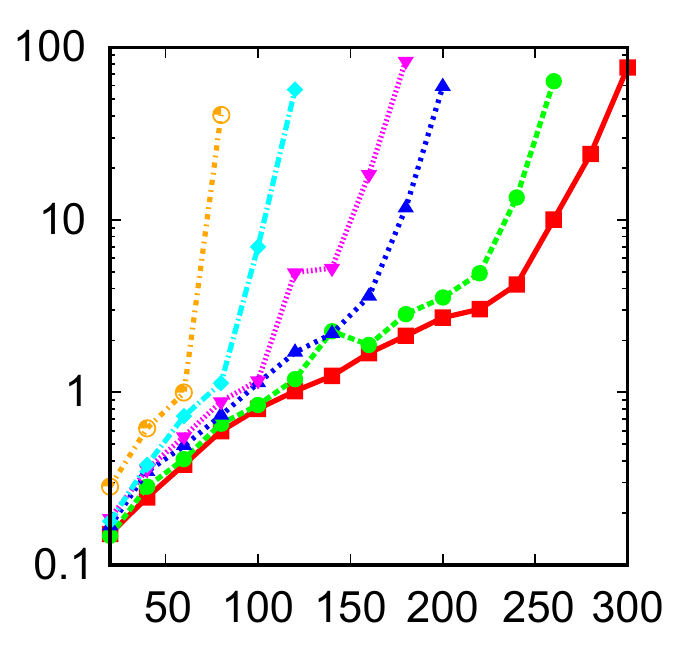} & 
		\includegraphics[width=1.5in]{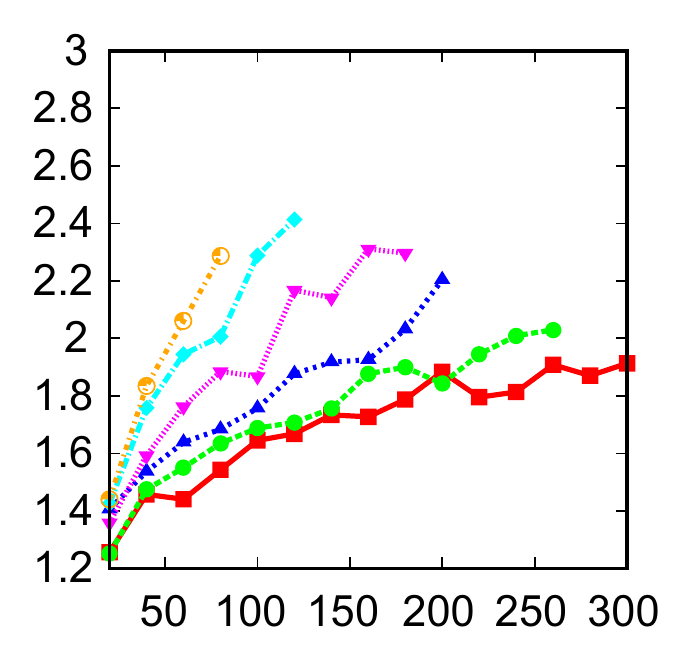}\vspace{1mm} \\
  \end{tabular}
\end{center}
\vspace*{-2mm}
\caption{Performance of the \tompp\, algorithm with $k$-way split for $k = 2$ (top row), $8$ (middle row), and $16$ (bottom row). The computation time and optimality ratio plots for each $k$ are shown side-by-side. The $x$-axis for all plots represents the number of robots. The $y$-axis for the plots on the left represents the computation time in seconds. The $y$-axis for the plots on the right represents the optimality ratio. The keys for all plots are the obstacle percentage, identical to that of Fig.~\ref{fig:24x18-ms}.} \label{fig:24x18-rest}
\end{figure}

The rest of the $k$-way split evaluation is presented in Fig.~\ref{fig:24x18-rest}, in which the computation time and optimality ratio are shown side by side without the axis labels. We also omit the key of the plots, which is the same as those from Fig.~\ref{fig:24x18-ms}, representing different obstacle percentages. In conjunction with Fig.~\ref{fig:24x18-ms} and Fig.~\ref{fig:24x18-ms-4}, as $k$ increases, we observe a general trend of reduced computation time at the expense of some optimality loss. With $16$-split, \tompp\, can solve problems with $300$ robots, corresponding to a robot-vertex ratio of $69\%$. 

\begin{figure}[htp]
\begin{center}
  \begin{tabular}{cc}
    \includegraphics[width=1.5in]{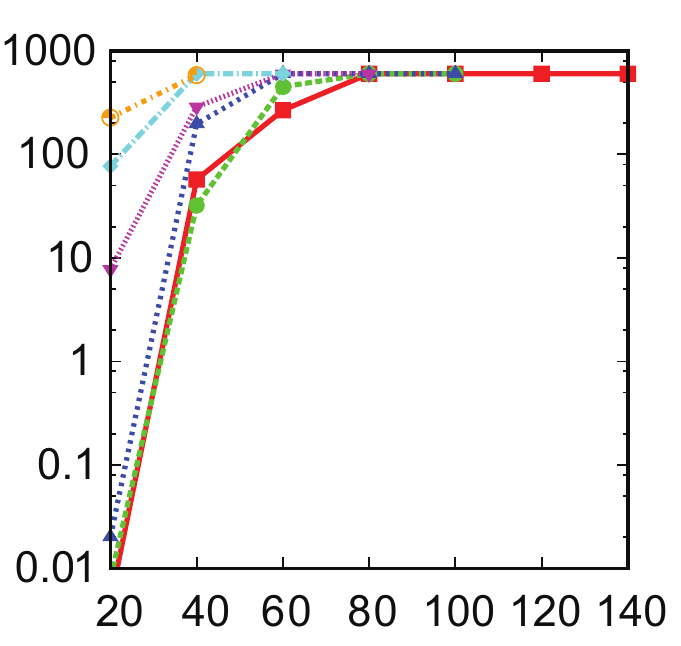} & 
		\includegraphics[width=1.5in]{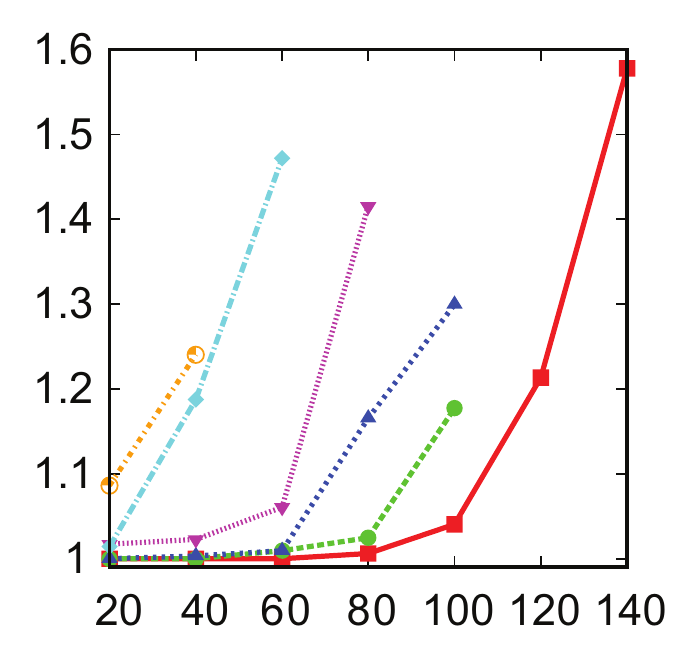} 
  \end{tabular}
\end{center}
\vspace*{-2mm}
\caption{Performance of the \ida-based anytime algorithm over the same set of problem instances. The plot setup is the same as that from Fig.~\ref{fig:24x18-rest}.} \label{fig:id-24x18-ms}
\end{figure}

For comparison, we ran \ida-based anytime algorithm over the same set of instances with a $600$-second time limit (we also attempted \odid\, and \wcha, which do not go past $40$ robots under the same setup). The result is plotted in Fig.~\ref{fig:id-24x18-ms}. \ida\, actually performs quite well for up to $60$ robots, which can be attributed to its A* root with minimum overhead as compared to our method. However, the performance of \ida\, degrades faster--it does not scale well beyond $100$ robots in our tests. \tompp, with $2$-way split, readily outperforms \ida\, when there are $40$ or more robots. Conceivably, it may be possible to combine \ida\, and $k$-way split to make it run faster. However, adding $k$-way split to \ida\, will inevitably make the overall makespan more sub-optimal. 

\subsection{Minimum Makespan Solution to $N^2$-puzzles}
Next, we evaluate the efficiency of the algorithm \tompp\, for finding minimum makespan solutions to the $N^2$-puzzle for $n=3, 4, 5,$ and $6$. These problems have a robot-vertex ratio of $100\%$, making them highly constrained and extremely challenging. Note that an $N^2$-puzzle instance is always solvable for $n \ge 3$ (see the Appendix); this means that all states are connected in the state (search) space. We ran Algorithm \tompp\, on 100 randomly generated $N^2$-puzzle instances for $n = 3, 4, 5$. For the 9-puzzle, computation on all instances completed successfully with an average computation time of 0.46 seconds per instance. To compare the computational result, we implemented a (optimal) BFS algorithm. The BFS algorithm is heavily optimized: For example, cycles of the grid are precomputed and hard coded to save computation time. Since the state space of the 9-puzzle is small, the BFS algorithm is capable of optimally solving the same set of 9-puzzle instances with an average computation time of about 1.08 seconds per instance. 

Once we move to the 16-puzzle, the power of general ILP solvers becomes evident. \tompp\, solved all 100 randomly generated 16-puzzle instances with an average computation time of 4.2 seconds. On the other hand, the BFS algorithm with a priority queue that worked for the 9-puzzle ran out of memory after a few minutes. As our result shows that an optimal solution for the 16-puzzle generally requires 6 time steps, it seems natural to also try bidirectional search, which cuts down the total number of states stored in memory. To complete such a search, one side of the bidirectional search generally must reach a depth of 3, which requires storing over $5 \times 10^8$ states (the branching factor is over 1000), each taking 64 bits of memory. This translates into over 4GB of raw memory and over 8GB of working memory, which is more than the JavaVM can handle: A bidirectional search ran out of memory after about 10 minutes in general. We also experimented with C++ implementation using STL libraries, which yields similar result ({\em i.e.}, ran out of memory before reaching a search depth of 3).

\begin{figure}[htp]
\begin{center}
    \includegraphics[width=0.16\textwidth]{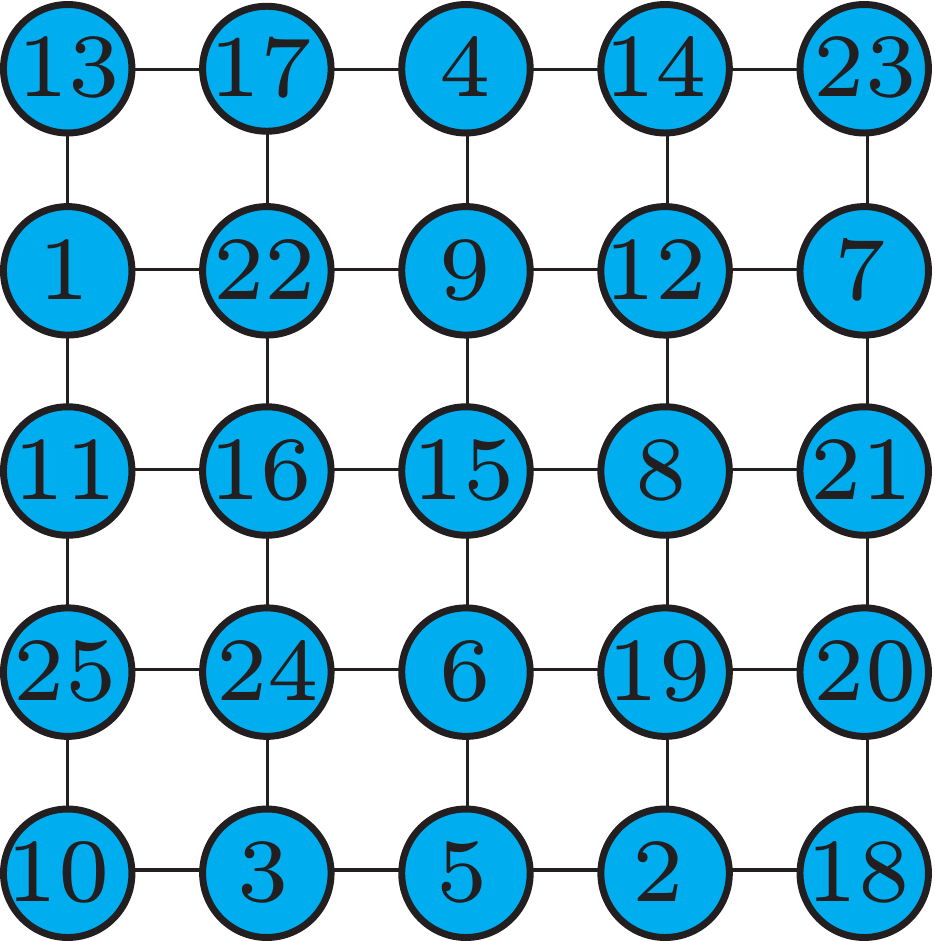} 
\end{center}
\vspace*{-2mm}
\caption{\label{fig:25-puzzle} An instance of a 25-puzzle problem solved by \tompp.}
\end{figure}

\begin{figure}[htp]
\begin{center}
    \includegraphics[width=0.45\textwidth]{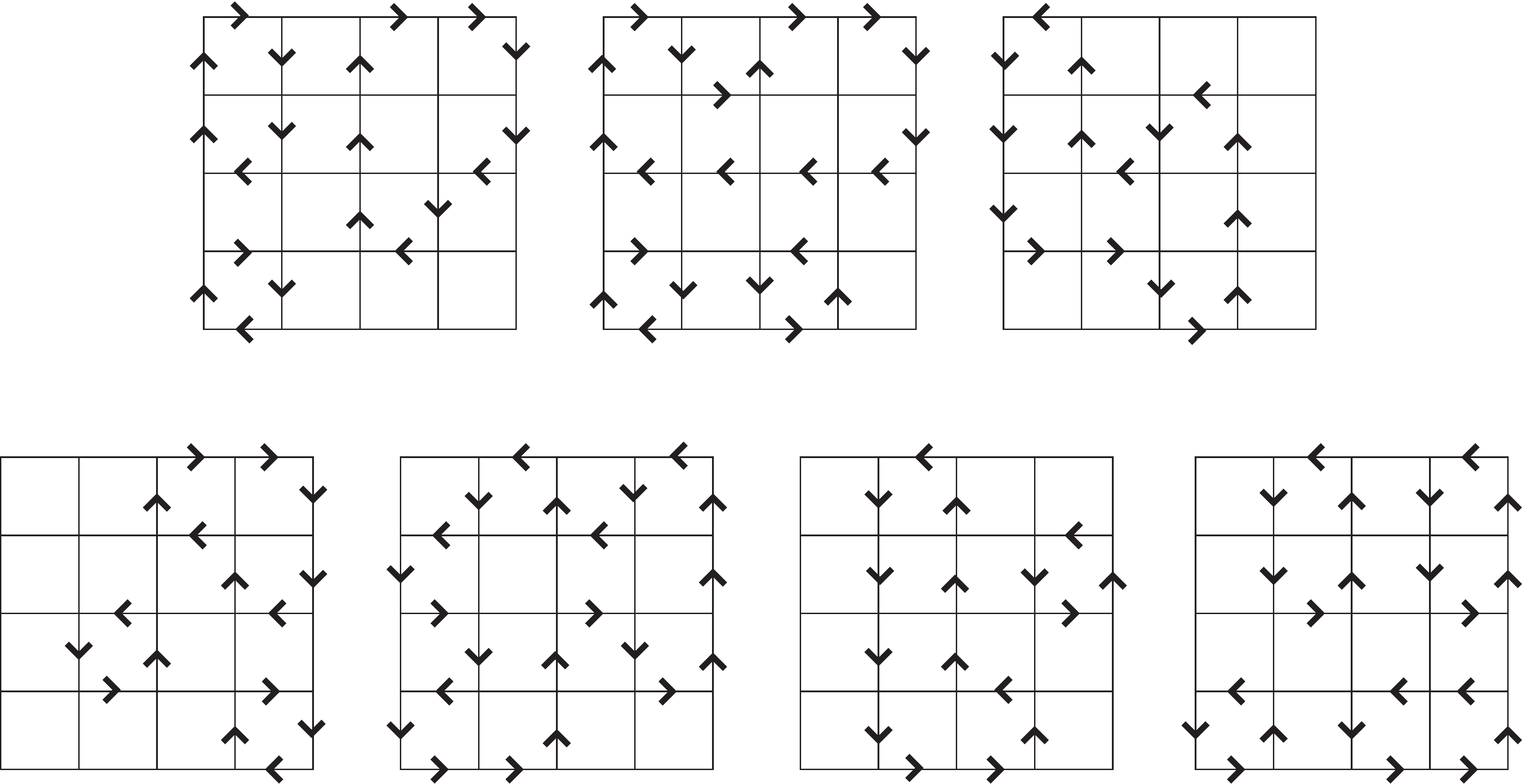} 
\end{center}
\caption{\label{fig:25-puzzle-sol} An optimal 7-step solution (from left to right, then top to bottom) to the 25-puzzle problem from Fig. \ref{fig:25-puzzle}, by \tompp\, in about $15$ minutes.}
\end{figure}

For the 25-puzzle, without a good heuristic, bidirectional search cannot explore a tiny fraction of the fully connected state space with about $10^{25}$ states. On the other hand, \tompp\, again consistently solves the 25-puzzle, with an average computational time of 391.6 seconds over 100 randomly created problems. Fig. \ref{fig:25-puzzle} shows one of the solved instances with a 7-step solution given in Fig. \ref{fig:25-puzzle-sol}. Note that 7 steps is obviously the least 
possible since it takes at least 7 steps to move robot 10 to its desired goal. We also briefly tested \tompp\, on the 36-puzzle. While we had some success here, \tompp\, generally does not seem to solve a randomly generated instance of the 36-puzzle within 24 hours, which has $3.7 \times 10^{41}$ states and a branching factor of well over $10^6$. 

As  comparison, \wcha\, can not solve $9$-puzzle. \odid, and \ida\, and can only solve $9$-puzzle consistently and cannot solve any $16$-puzzles in $600$ seconds. 

\subsection{Minimum Makespan on $8\times 8$, $16\times 16$, and $32\times 32$ Grids}
In this section, we evaluate \tompp\, with underlying graphs that are $8 \times 8$ grids, $16\times 16$ grids, and $32 \times 32$ grids with $20\%$ obstacles. In addition to further demonstrating the effectiveness of \tompp, this allows us to better compare our results. $8 \times 8$ and $16\times 16$ grids are used as the environment for evaluation in \cite{Sur12}. $32 \times 32$ grids with $20\%$ obstacles are used for evaluation in \cite{Sil05,StaKor11}. 

\begin{figure}[htp]
\begin{center}
  \begin{tabular}{cc}
    \includegraphics[width=1.5in]{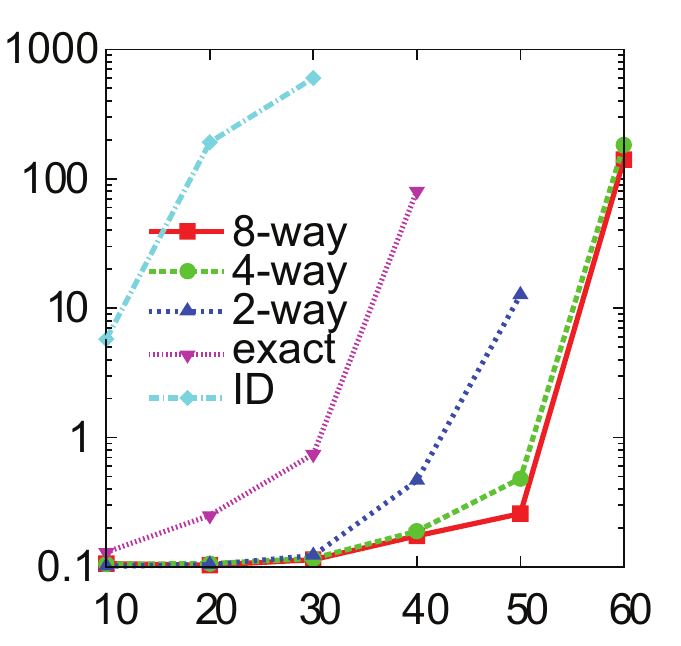} & 
		\includegraphics[width=1.5in]{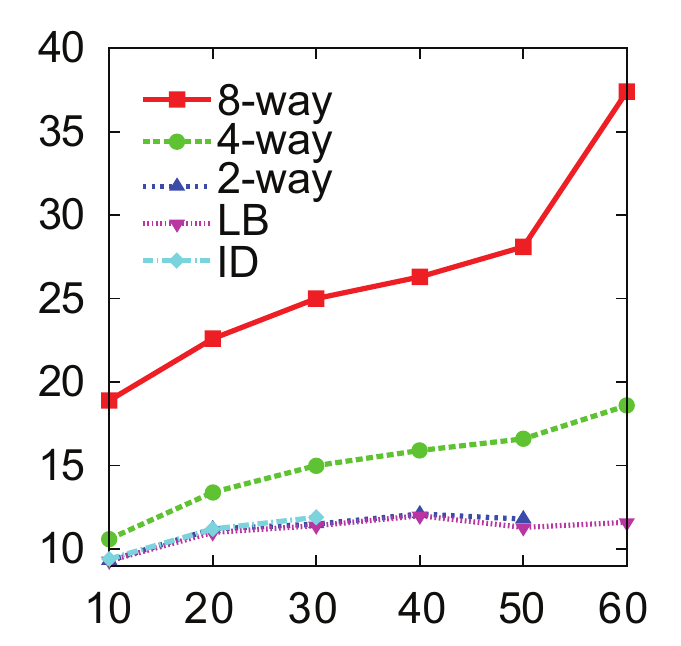} \vspace{1mm}\\
    \includegraphics[width=1.5in]{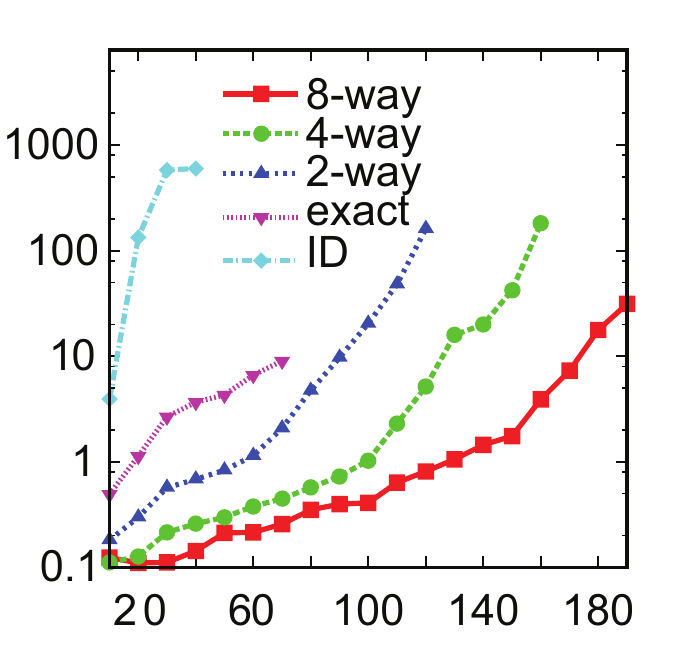} & 
		\includegraphics[width=1.5in]{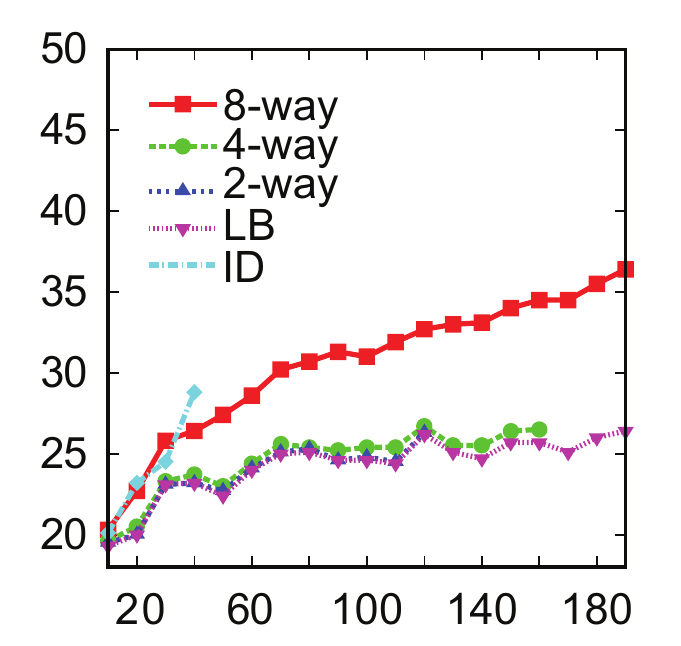} 
  \end{tabular}
\end{center}
\vspace*{-2mm}
\caption{Performance of the \tompp\, algorithm with $k$-way split for $k = 2$-$8$ and the \ida-based anytime algorithm. The computation time and solution makespan plots are shown side-by-side. The $x$-axis for all plots represents the number of robots. The $y$-axis for the plots on the left represents the computation time in seconds. The $y$-axis for the plots on the right represents the solution makespan. Note that we did not plot the makespan computed by the exact \tompp\, algorithm. Instead, the lower bound (LB) estimate of makespan is plotted (in magenta color). [top] Result on the $8\times 8$ grid. [bottom] Result on the $16\times 16$ grid.} \label{fig:16x16}
\end{figure}

For $8\times 8$ and $16\times 16$ grids, the instances are constructed using the same procedure stated in Section~\ref{subsection:k-way-split}. Again, each data point is an average over $10$ sequentially randomly created instances. Given the size of $8\times 8$ and $16 \times 16$ grids, we limit $k$ to $8$; using $16$-way split can solve more instances but incur an average solution optimality between $2$-$4$ optimal. Each instance is given a time limit of $600$ seconds. The outcome of these experiments is plotted in Fig.~\ref{fig:16x16}, along with the result from running \ida. \odid\, and \wcha\, cannot consistently solve the instances with $20$ robots in $600$ seconds. Makespan, instead of optimality ratio, is used in Fig.~\ref{fig:16x16} for easy comparison with the results from \cite{Sur12}. 

We observe that, over the $8\times 8$ grid, with $2$-way split heuristic, \tompp\, can solve problems with $50$ robots to almost true optimal solutions in just $10$ seconds. With $4$-way split, \tompp\, can further push to $60$ robots (robot-vertex ratio of $94\%$) with solutions that are within $1.7$-optimal. \ida\, can only handle up to $30$ robots. As reported in \cite{Sur12}, \cobopt\, generally takes more than half an hour to produce its final solution when there are $24$ or more robots.\footnote{We did not directly run \cobopt\, over our randomly created instances. However, the instances in \cite{Sur12} are created in an identical manner. Therefore, given that similarly powered computers are used, the computation time and solution makespan are directly comparable between ours and those from \cite{Sur12}.} The solution quality also degrades quickly as the number of robots increases. For example (Fig. 2 and Fig. 3 in \cite{Sur12}), at $50$ robots, \cobopt\, takes over an hour to produce a solution with a makespan of over $160$, whereas our $2$-way split heuristic yields a near-optimal makespan of $11.8$ in just $10$ seconds. 

Over the $16 \times 16$ grid, \tompp\, is able to handle instances with $190$ (robot-vertex ratio of $74\%$) robots with $8$-way split while at the same yielding solutions that are always less than $1.4$-optimal. When switched to $4$-way split, \tompp\, can consistently solve problems with up to $160$ robots to no more than $1.03$-optimal. In comparison, \ida\, can solve instances with up to $80$ robots to relatively good quality. Taking on average an hour of computation, \cobopt\, can handle up to $128$ robots; the solution quality is quite poor. For example (Fig. 4 and Fig. 5 in \cite{Sur12}), for about $100$ robots, the computed makespan by \cobopt\, is at $200$ whereas the optimal makespan is about $25$. Across $8\times 8$ and $16\times 16$ grids, we generally observe a speedup of over $100$ times when \tompp\, (with the $4$-way split heuristic) is compared with \cobopt. At the same time, our method yields solutions with much smaller makespan. 

\begin{figure}[htp]
\begin{center}
  \begin{tabular}{cc}
    \includegraphics[width=1.5in]{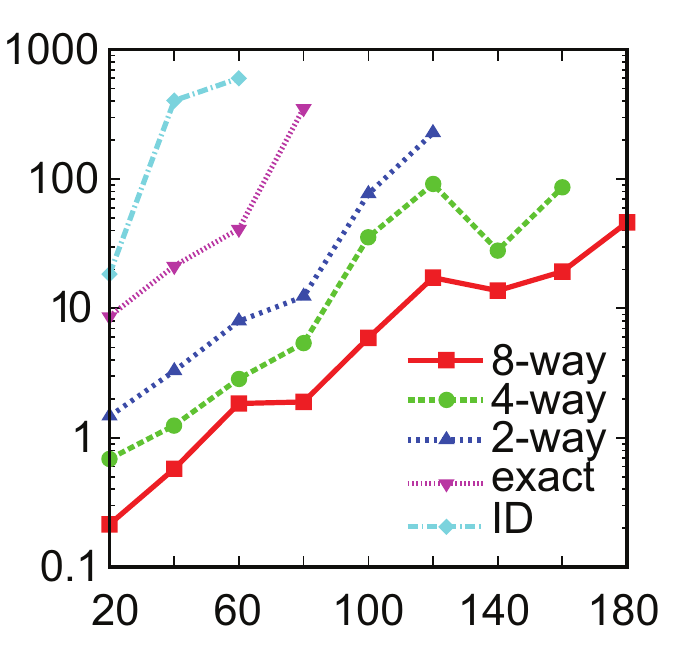} & 
		\includegraphics[width=1.5in]{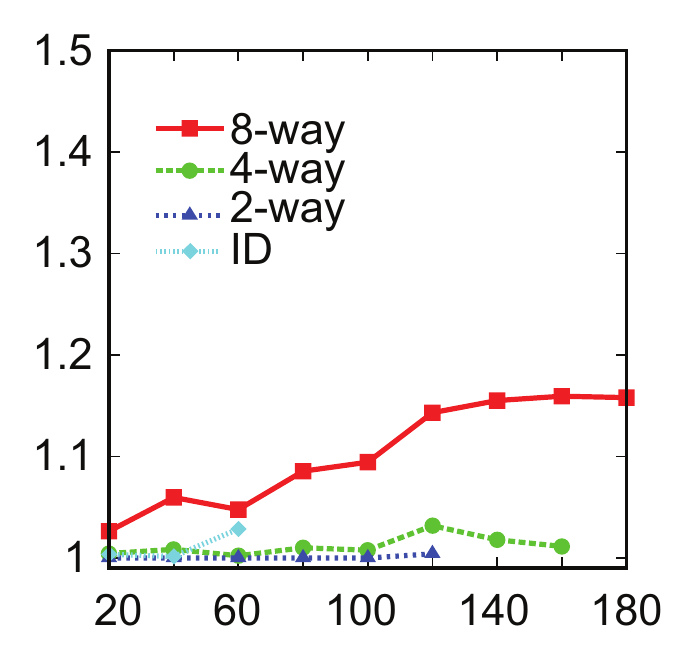} 
  \end{tabular}
\end{center}
\vspace*{-2mm}
\caption{Performance of \tompp\, and the \ida-based anytime algorithm over $32\times 32$ grid. The axis setup is the same as that from Fig.~\ref{fig:24x18-rest}.} \label{fig:id-32x32-ms}
\end{figure}

A classical test scenario is the $4$-connected $32\times 32$ grid with $20\%$ vertices randomly removed.\footnote{Some work ({\em e.g.}, \cite{StaKor11}) also adopts an $8$-connected model. That is, each vertex is on the grid is connected to its $8$-neighborhood. This causes unit cost to be assigned to all edges, although a diagonal edge should have length $\sqrt{2}$ times that of a non-diagonal edge. Since we are modeling robots in this work, we do not discuss the $8$-connected model here. However, we mention that our algorithms easily extend to $8$-connected model. Our tests show that we can in fact compute near-optimal makespan for $400$ robots on $32 \times 32$ grids assuming $8$-connectivity.} For completeness, we also perform a brief evaluation of this setup. We randomly generate the instance as before, run the test, and plot the result in Fig.~\ref{fig:id-32x32-ms}. From the figure, we observe a pattern consistent with experiments on the $8\times 8$, $16\times 16$ and $24\times 18$ grids. 

\subsection{Algorithm Performance over All Objectives}
Last, we evaluate the performance of our algorithms at optimizing all Objectives~\ref{omakespan}-\ref{otd}. The result on \tompp\, is already presented in Section~\ref{subsection:k-way-split}, which we also use as the solution to optimizing Objective~\ref{md} ({\em i.e.}, we simply use \tompp\, in place of \mdmpp\, due to its superior performance). Note that no change to the plots are needed here because, on one hand, we can use a (near)-optimal makespan solution as a solution to minimizing maximum single-robot traveled distance. This is true because the makespan of a solution is always no less than the minimum maximum single-robot distance. On the other hand, the lower bound estimate for minimum makespan is the same as that for minimum maximum single-robot distance.

Before moving to \ttmpp\, and \dompp, we note that these algorithms possess some properties of an {\em anytime algorithm}, which is of practical importance. In solving these ILP models, the solver generally uses variations of the {\em branch-and-bound} algorithm \cite{LanDoi60}. For computing total time (and distance) optimal solutions, a branch-and-bound algorithm always try to find a feasible solution first and then iteratively improve over the feasible solution. This naturally leads to improving solution quality commonly observed in an anytime algorithm. The anytime property of \ttmpp\, and \dompp\, allows us to set a desired sub-optimal threshold to reduce the computation time. Note that the same cannot be said for \tompp\, because a feasible solution here is the optimal solution. 

\begin{figure}[htp]
\begin{center}
  \begin{tabular}{c}
    \includegraphics[width=2.8in]{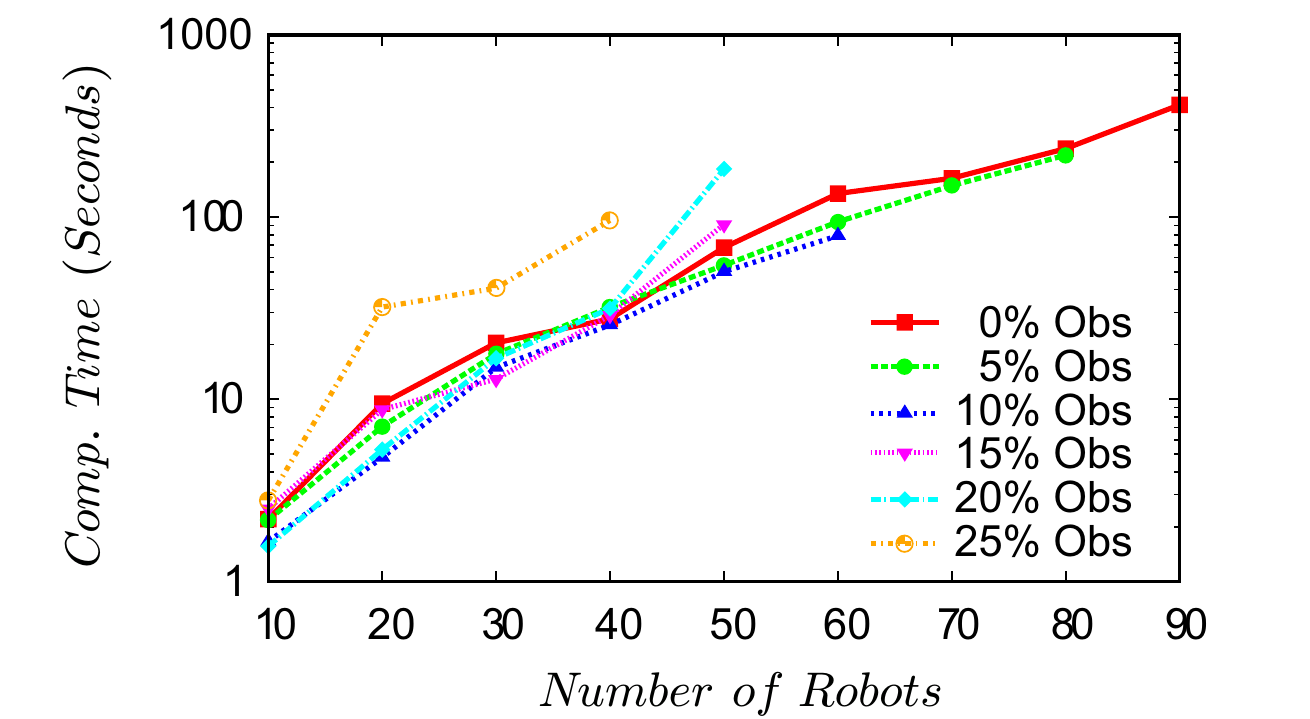} \vspace{1mm}\\
		\includegraphics[width=2.8in]{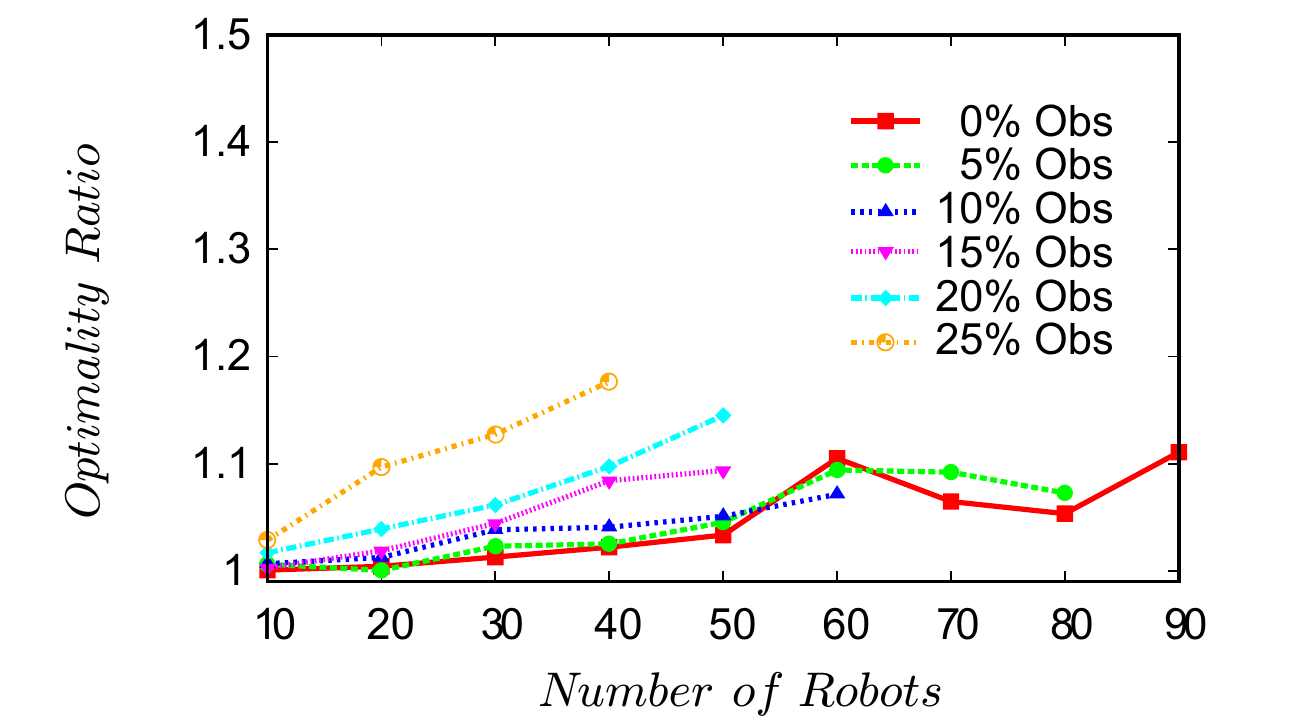}
  \end{tabular}
\end{center}
\vspace*{-2mm}
\caption{[top] Average computation time of \ttmpp\, algorithm over instances on a $24 \times 18$ grid with randomly placed obstacles and start/goal locations. [bottom] The achieved (conservatively estimated) optimality ratio. } \label{fig:tt-plot}
\end{figure}

Our next set of results focuses on the \ttmpp\, algorithm (Fig.~\ref{fig:tt-plot}). The general setup is the same as that used in Section~\ref{subsection:k-way-split}. In particular, the same set of problem instances is used. In out experiment, we limit both time ($600$ seconds) and required sub-optimality threshold (automatically adjusted) to achieve a balanced performance. The lower bound estimate for computing optimality ratio is obtained by summing over the individual shortest path lengths. The actual optimal total time is about $15$ times the number of robots ({\em i.e.}, $150$ for $10$ robots and $1500$ for $100$ robots), regardless of the percentage of obstacles. We observe that the \ttmpp\, algorithm is fairly effective, capable of computing $1.1$-optimal solutions for up to $100$ robots in the allocated time. 

\begin{figure}[htp]
\begin{center}
  \begin{tabular}{c}
    \includegraphics[width=2.8in]{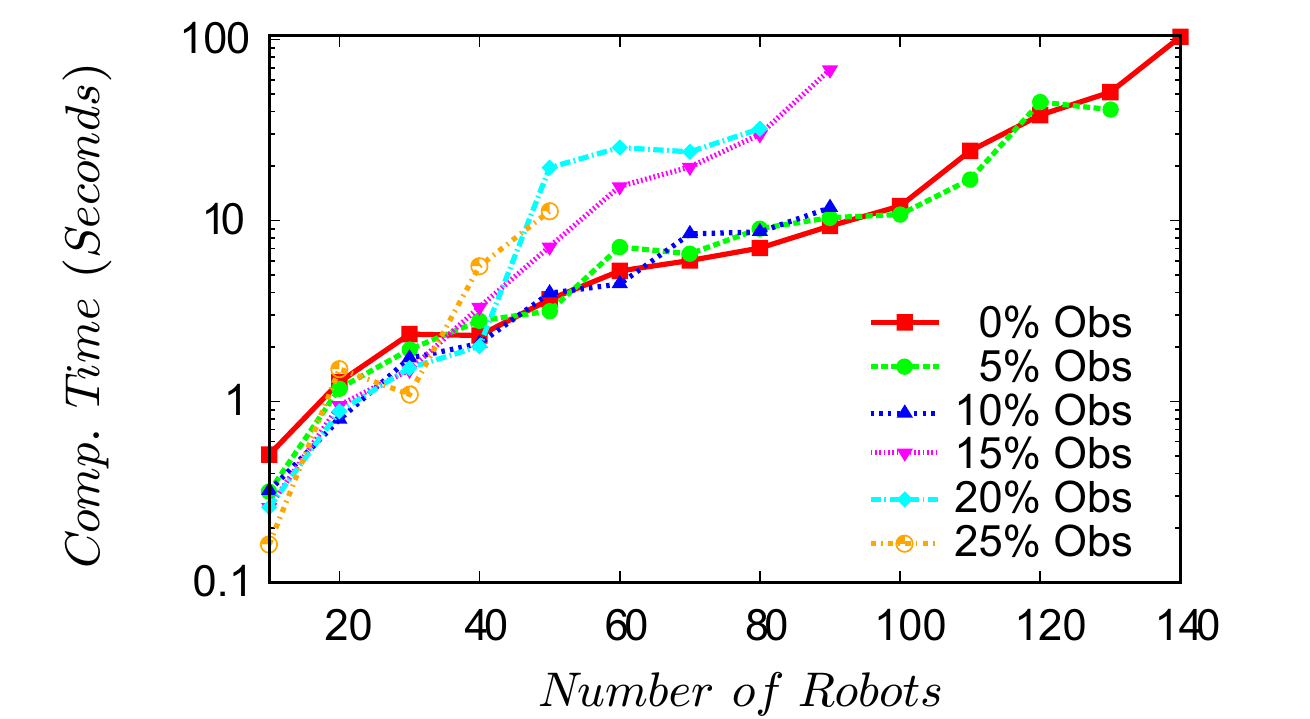} \vspace{1mm}\\
		\includegraphics[width=2.8in]{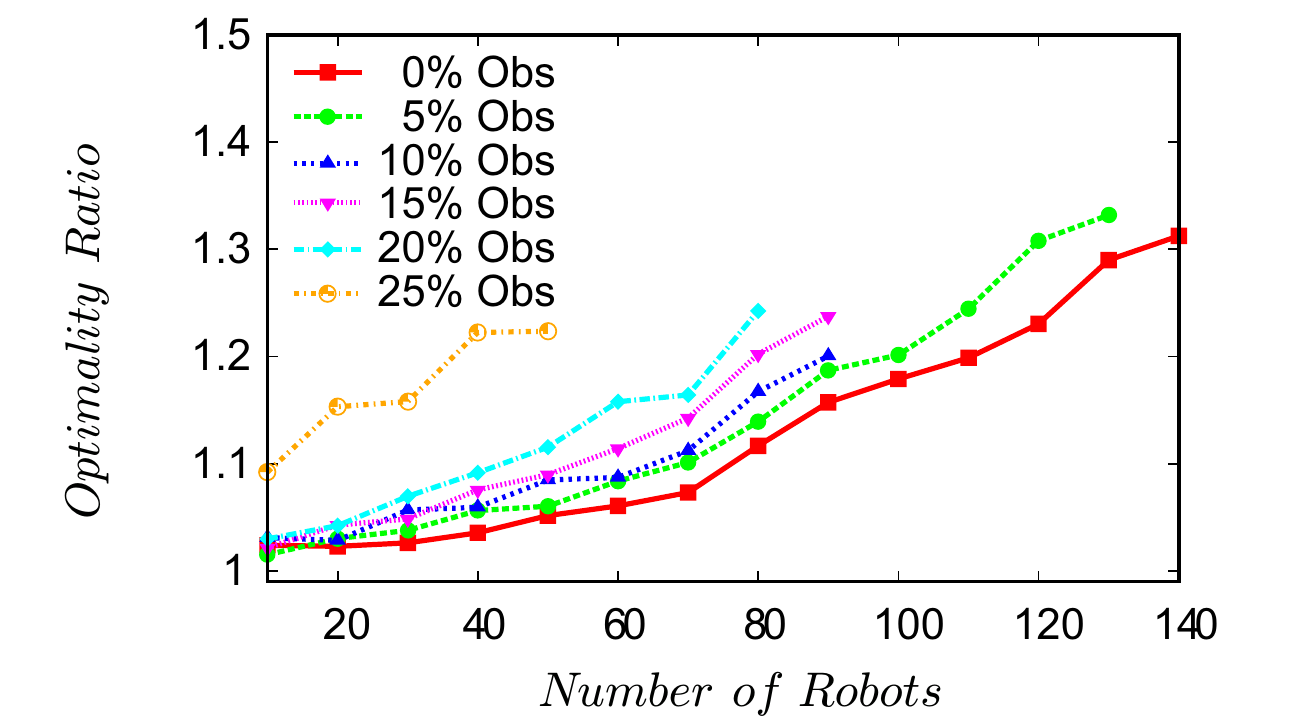}
  \end{tabular}
\end{center}
\vspace*{-2mm}
\caption{[top] Average computation time of \dompp\, algorithm over instances on a $24 \times 18$ grid with randomly placed obstacles and start/goal locations. [bottom] The achieved (conservatively estimated) optimality ratio. } \label{fig:td-plot}
\end{figure}

Similar outcome is also observed in the performance evaluation of the \dompp\, algorithm with the $4$-way split heuristic (Fig.~\ref{fig:td-plot}). The optimal total distance is again about $15$ times the number of robots. In comparison with the total time optimal case, due to the $4$-way split heuristic, the \dompp\, algorithm is faster but produced solutions that are more sub-optimal but still quite good. 

\begin{figure}[htp]
\begin{center}
  \begin{tabular}{cc}
    \includegraphics[width=1.5in]{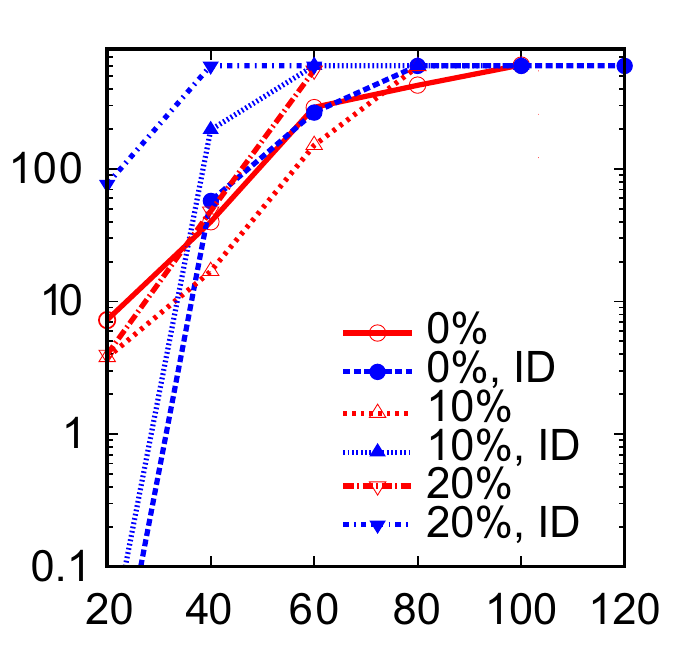} & 
		\includegraphics[width=1.5in]{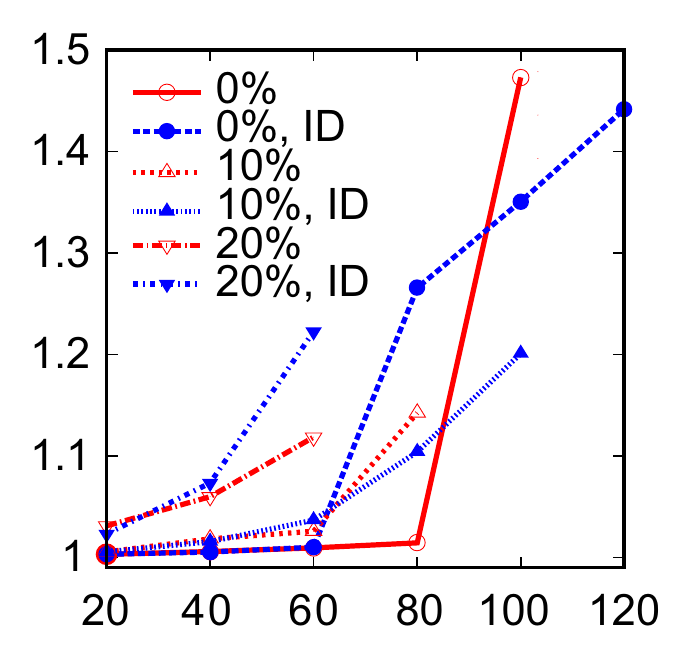} \vspace{1mm} \\
    \includegraphics[width=1.5in]{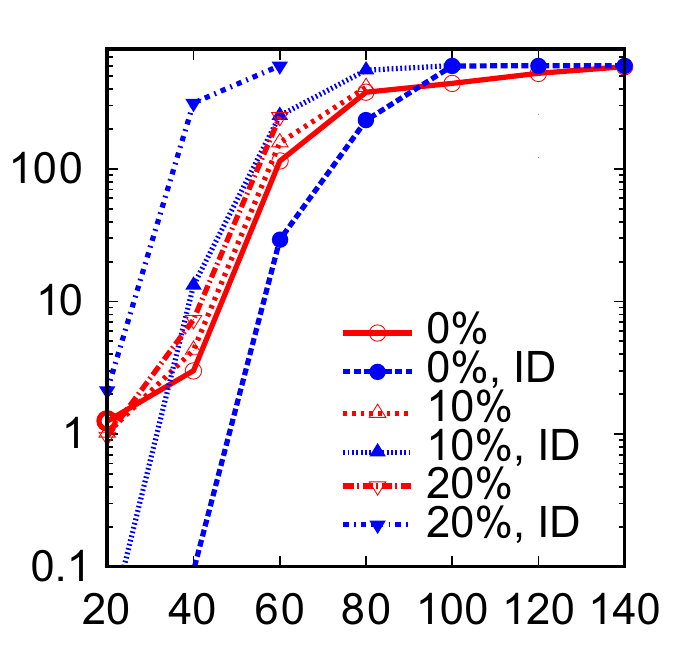} & 
		\includegraphics[width=1.5in]{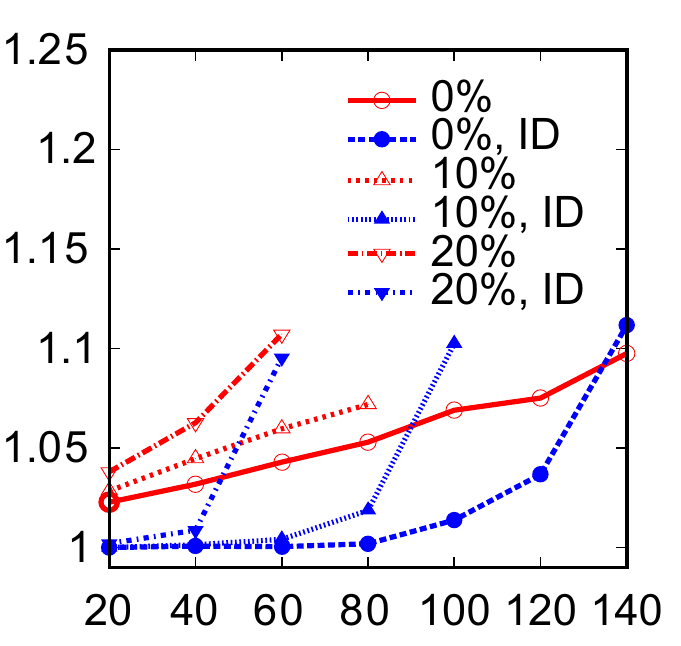}  \\
  \end{tabular}
\end{center}
\vspace*{-2mm}
\caption{Performance of \ttmpp\, \dompp, and the \ida-based anytime algorithm over $24\times 18$ with $0$-$20\%$ obstacles. Each instance is allowed $600$ seconds of time. The axis setup is the same as that from Fig.~\ref{fig:24x18-rest}. [top] \ttmpp\, versus total time \ida; the red lines correspond to data for \ttmpp. [bottom] \dompp\, versus total distance \ida; the red lines correspond to data for \dompp.} \label{fig:idc}
\end{figure}

For comparison, we run \ttmpp, \dompp, and \ida\,(total time and total distance versions) on $24 \times 18$ grids with $0$-$20\%$ obstacles with a maximum time limit of $600$ seconds. The setting is slightly different from that used in obtaining Fig.~\ref{fig:tt-plot} and~\ref{fig:td-plot}; we do not set a sub-optimal threshold here. The result is plotted in Fig.~\ref{fig:idc}. In the case of total time (Objective~\ref{ott}), \ida\, could solve more instances. For the instances that are solved, the achieved optimality is similar. For total distance (Objective~\ref{otd}), we observe similar outcome. Here, \ida\, could produce one more data point; the achieved optimality by both methods are again comparable (and good). Overall, \ttmpp, and \dompp\, are competitive with \ida. 

%\remark We point out that \ttmpp\, and \dompp\, algorithms possess some properties of an {\em anytime algorithm}, which is of practical importance. In solving these ILP models, the solver generally uses variations of the {\em branch-and-bound} algorithm \cite{LanDoi60}. For computing total time (and distance) optimal solutions, a branch-and-bound algorithm always try to find a feasible solution first and then iteratively improve over the feasible solution. This naturally leads to improving solution quality commonly observed in an anytime algorithm. Note that the same cannot be said for \tompp\, because a feasible solution here is the optimal solution. ~\rqed

\section{Conclusion}\label{sec:conclusion}
In this paper, we propose a general algorithmic framework for solving $\mpp$ problems optimally or near-optimally. Through an equivalence relationship between $\mpp$ and network flow, we provide ILP model-based algorithms for minimizing the makespan (last arrival time), the maximum (single-robot traveled) distance, the total arrival time, and the total distance. In conjunction with additional heuristics, our algorithmic solutions are highly effective, capable of computing near-optimal solutions for hundreds of robots in seconds in scenarios with very high robot-vertex ratio. In pushing for high performance algorithms aim at solving $\mpp$ optimally or near-optimally, our eventual goal is to apply these algorithms to multi-robot path planning problems in continuous domains--our preliminary work toward this goal has begun to show promising results, producing algorithms that can compute solutions for around a hundred disc robots in 2D environments with holes. 

%Many open questions remain, we mention two here. To solve larger problems instances, \dompp\, and \ttmpp\, still require non trivial amount of time. This prevents them from being used as a subroutine in time critical systems. Additional refinements are needed to make these algorithms more efficient, which may be possible by further simplifying the models. 

\bibliographystyle{IEEEtranN}
\bibliography{jingjin}

\section*{Appendix}

\subsection{Properties of the $N^2$-puzzle}\label{sec:puzzle}
The example problem from Fig. \ref{fig:example} easily extends to an $N \times N$ grid; we call this class of problems the $N^2$-puzzle. Such problems are highly coupled: No robot can move without at least three other robots moving at the same time. At each step, all robots that move must move synchronously in the same direction (per cycle) on one or more disjoint cycles (see {\em e.g.}, Fig. \ref{fig:puzzle-8-sol}). To put into perspective the computational results on $N^2$-puzzles that follow, we make a characterization of the state structure of the $N^2$-puzzle for $N \ge 3$ (the case of $N=2$ is trivial). 
\begin{figure}[htp]
\begin{center}
    \includegraphics[width=0.45\textwidth]{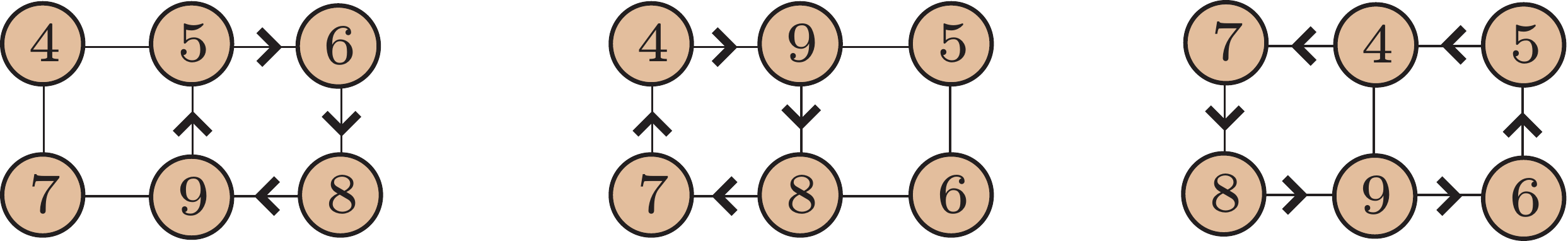} 
\end{center}
\vspace*{-2mm}
\caption{\label{fig:6-puzzle} A 3-step procedure for exchanging robots 8 and 9.}
\end{figure}

\begin{proposition}\label{p:state}All states of the 9-puzzle are connected via legal moves. 
\end{proposition}
{\sc Proof}. We show that any state of a 9-puzzle can be moved into the state shown in Fig. \ref{fig:example}(b). From any state, robot 5 can be easily moved into the center of the grid. We are left to show that we can exchange two robots on the border without affecting other robots. This is possible due to the procedure illustrated in Fig. \ref{fig:6-puzzle}. 
~\qed

\begin{figure}[htp]
\begin{center}
    \includegraphics[width=0.45\textwidth]{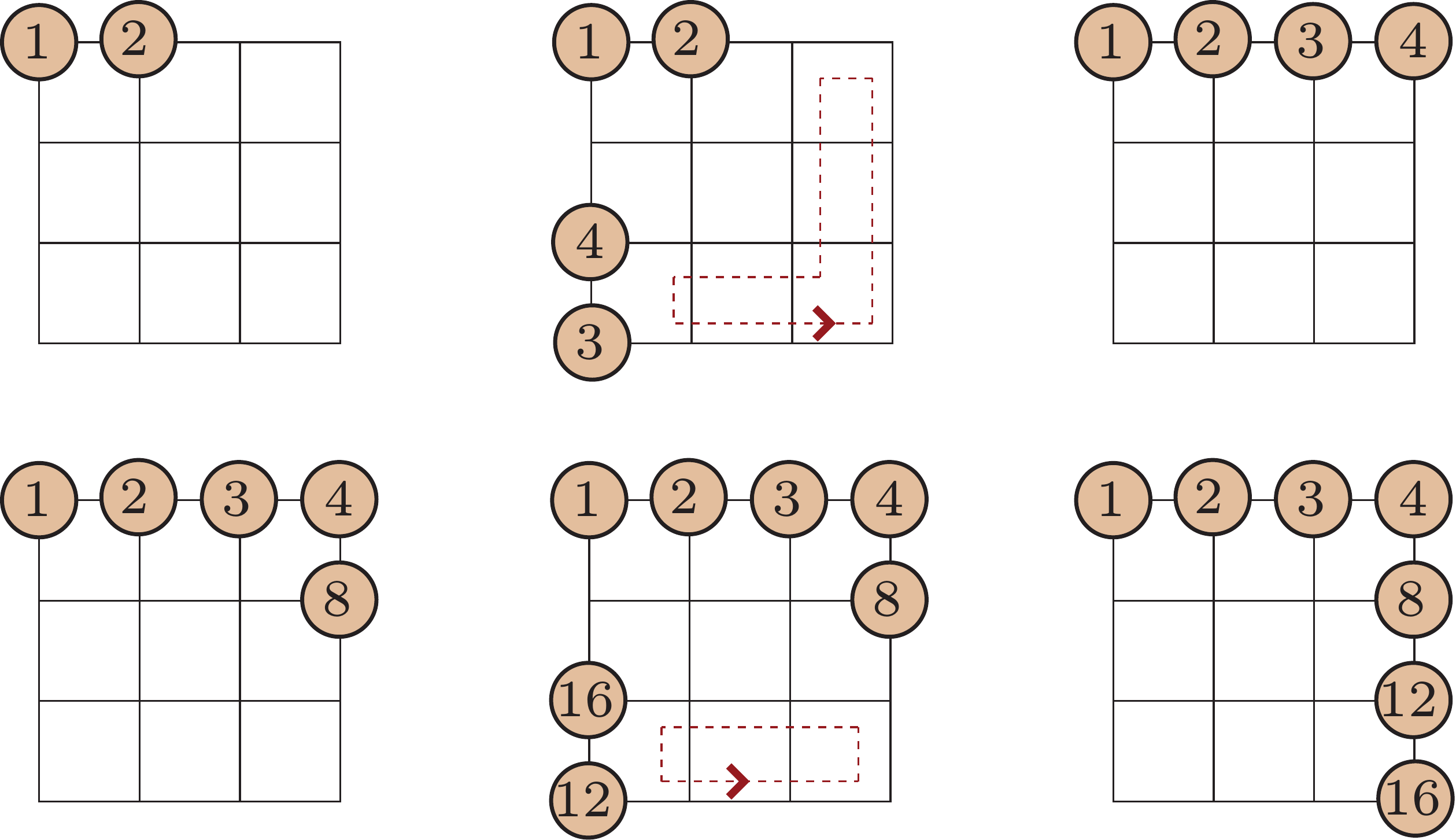} 
\end{center}
\vspace*{-2mm}
\caption{\label{fig:16-puzzle} A solution scheme for solving top/left sides of the 16-puzzle.}
\end{figure}

Larger puzzles can be solved recursively: We may first solve the top and right side of the puzzle and then the left over smaller square puzzle. For a 16-puzzle, Fig. \ref{fig:16-puzzle} outlines the procedure, consisting of six main steps:
\begin{enumerate}
\item Move robots 1 and 2 to their respective goal locations, one robot at a time (first 1, then 2). 
\item Move robots 3 and 4 (first 3, then 4) to the lower left corner (top-middle figure in Fig. \ref{fig:16-puzzle}). 
\item Move robots 3 and 4 to their goal location together via counterclockwise rotation along the cycle indicated in the top-middle figure in Fig. \ref{fig:16-puzzle}. 
\item Move robot 8 to its goal location. 
\item Move robots 12 and then 16 to the lower left corner. 
\item Rotate robots 12 and 16 to their goal locations. 
\end{enumerate}

It is straightforward to see that larger puzzles can be solved similarly. We have thus outlined the essential steps for proving Proposition \ref{c:state} below; a more generic proof can be written using generators of permutation groups, which we omit here due to its length. Proposition \ref{c:state} implies that, for $N \ge 3$, all instances of $N^2$-puzzles are solvable. The constructive proofs of Proposition \ref{p:state} and \ref{c:state} lead to recursive algorithms for solving any $N^2$-puzzle (clearly, the solution is not time/distance optimal in general). 

\begin{theorem}\label{c:state}All states of an $N^2$-puzzle, $N \ge 3$ are connected via legal moves. 
\end{theorem}
\begin{corollary}\label{c:solvable}All instances of the $N^2$-puzzle, $N \ge 3$, are solvable. 
\end{corollary}

By Proposition \ref{c:state}, since all states of a $N^2$-puzzle for $N \ge 3$ are connected via legal moves, the state space of searching an $N^2$-puzzle equals $N^2$ {\em factorial}. For  16-puzzle and 25-puzzle, $16! > 10^{13}, 25! > 10^{25}$. Large state space is one of the three reasons that make finding a time optimal solution to the $N^2$-puzzle a difficult problem. The second difficulty comes from the large branching factor at each step. For a 9-puzzle, there are 13 unique cycles, yielding a branching factor of 26 (clockwise and counterclockwise rotations). For the 16-puzzle, the branching factor is around 500. This number balloons to over $10^4$ for the 25-puzzle. This suggests that on typical commodity personal computer hardware (assuming a 1GHz processor), a baisc breadth first search algorithm will not be able to go beyond depth of 3 for the 16-puzzle and depth 2 for the 25-puzzle in reasonable amount of time. Moreover, enumerating these cycles is a non-trivial task. The third difficulty is the lack of obvious heuristics: Manhattan distances of robots to their respective goals prove to be a bad one. For example, given the initial configuration as that in Fig. \ref{fig:example}(a), the first step in the optimal plan from Fig. \ref{fig:puzzle-8-sol} gets robots 1, 3, 4, 6, 8, 9 closer to their respective goals while moving robots 2, 7 farther. On the other hand, rotating counterclockwise along the outer cycle takes robots 1, 3, 4, 5, 6, 8, 9 closer and only moves robot 7 farther. However, if we instead take this latter first step, the optimal plan afterward will take 5 more steps. 

\end{document}